%% file: main.tex
\documentclass{article}

\PassOptionsToPackage{numbers}{natbib}

     \usepackage[final]{neurips_2020}

\usepackage[utf8]{inputenc} %
\usepackage[T1]{fontenc}    %
\usepackage{hyperref}       %
\usepackage{url}            %
\usepackage{booktabs}       %
\usepackage{amsfonts}       %
\usepackage{nicefrac}       %
\usepackage{microtype}      %

\usepackage{microtype}
\usepackage{graphicx}
\usepackage{subfigure}
\usepackage{booktabs}

\usepackage[utf8]{inputenc} %
\usepackage[T1]{fontenc}    %
\usepackage{hyperref}       %
\usepackage{url}            %
\usepackage{booktabs}       %
\usepackage{amsmath,amssymb,amsfonts,amsthm}       %
\usepackage{nicefrac}       %
\usepackage{microtype}      %
\usepackage{graphicx}
\usepackage{arydshln}
\usepackage{color}
\usepackage{dsfont}
\usepackage{caption}
\usepackage{appendix}
\usepackage{mathrsfs}
\input{macros}

\usepackage{hyperref}

\title{Reward-rational (implicit) choice:\\ A unifying formalism for reward learning}

\author{Hong Jun Jeon$^{*1}$, Smitha Milli$^{*2}$, Anca Dragan$^{2}$ \\
\texttt{hjjeon@stanford.edu, smilli@berkeley.edu, anca@berkeley.edu} \\ 
\small{$^{*}$Equal contribution}, \\
\small{$^{1}$Stanford University}, \\
\small{$^{2}$University of California, Berkeley}}

\begin{document}

\maketitle

\begin{abstract}
It is often difficult to hand-specify what the correct reward function is for a task, so researchers have instead aimed to learn reward functions from human behavior or feedback. The types of behavior interpreted as evidence of the reward function have expanded greatly in recent years. We've gone from demonstrations, to comparisons, to reading into the information leaked when the human is pushing the robot away or turning it off. And surely, there is more to come. How will a robot make sense of all these diverse types of behavior?
Our key observation is that different types of behavior can be interpreted in a single unifying formalism - as a \emph{reward-rational choice} that the human is making, often implicitly. We use this formalism to survey prior work through a unifying lens, and discuss its potential use as a recipe for interpreting new sources of information that are yet to be uncovered. %
\end{abstract}

\input{intro.tex}

\input{unifying_framework.tex}

\input{applying_formalism.tex}

\input{disc.tex}

\section*{Broader Impact}
As AI capability advances, it is becoming increasingly important to align the objectives of AI agents to what people want. From how assistive robots can best help their users, to how autonomous cars should trade off between safety risk and efficiency, to how recommender systems should balance revenue considerations with longer-term user happiness and with avoiding influencing user views, agents cannot rely on a reward function specified once and set in stone. By putting different sources of information about the reward explicitly under the same framework, we hope our paper contributes towards a future in which agents maintain uncertainty over what their reward should be, and use different types of feedback from humans to refine their estimate and become better aligned with what people want over time -- be them designers or end-users. 

On the flip side, changing reward functions also raises its own set of risks and challenges. First, the relationship between designer objectives and end-user objectives is not clear. Our framework can be used to adapt agents to end-users preferences, but this takes away control from the system designers. This might be desirable for, say, home robots, but not for safety-critical systems like autonomous cars, where designers might need to enforce certain constraints a-priori on the reward adaptation process. More broadly, most systems have multiple stake-holders, and what it means to do ethical preference aggregation remains an open problem. Further, if the robot's model of the human is misspecified, adaptation might lead to more harm than good, with the robot inferring a worse reward function than what a designer could specify by hand. 

\begin{ack}
We thank the members of the InterACT lab for fruitful discussion and advice, especially Dylan Hadfield-Menell for his perspectives on the relationship between demonstrations and comparisons. We thank Andreea Bobu, Paul Christiano, and Rohin Shah for their feedback on the manuscript. 

This work is partially supported by ONR YIP and Open Philanthropy Project. This material is based upon work supported by the National Science Foundation Graduate Research Fellowship under Grant No. 1752814. Any opinion, findings, and conclusions or recommendations expressed in this material are those of the authors(s) and do not necessarily reflect the views of the National Science Foundation.
\end{ack}

\bibliographystyle{unsrtnat}
\bibliography{refs}

\clearpage
\input{appendix.tex}

\end{document}

%% file: macros.tex
\newcommand{\R}{\mathbb{R}}

\newcommand{\pr}[0]{\mathbb{P}}
\newcommand{\trajs}[0]{\Xi}
\newcommand{\traj}[0]{\xi}
\newcommand{\choices}[0]{\mathcal{C}}

\newcommand{\reward}[0]{r}
\newcommand{\rewards}[0]{\mathcal{R}}
\newcommand{\map}[0]{\psi}

\newcommand{\fdbk}[0]{c}
\newcommand{\fdbkch}[0]{c^*}
\newcommand{\rob}[0]{R}

\newcommand{\rat}[0]{\beta}
\newcommand{\ex}[0]{\mathbb{E}}

\newcommand{\error}[0]{\epsilon}

\newcommand{\langset}[0]{\Lambda}
\newcommand{\lang}[0]{\lambda}
\newcommand{\lgrding}[0]{G}
\newcommand{\st}[0]{s}

\newcommand{\figref}[1]{Fig. \ref{#1}}
\newcommand{\sref}[1]{Sec. \ref{#1}}

\DeclareMathOperator*{\argmax}{arg\,max}
\DeclareMathOperator*{\argmin}{arg\,min}

\newtheorem{theorem}{Theorem}[section]
\newtheorem{cor}{Corollary}[section]
\theoremstyle{definition}
\newtheorem{definition}{Definition}[section]

%% file: intro.tex
\section{Introduction}
It is difficult to specify reward functions that always lead to the desired behavior. Recent work has argued that the reward specified by a human is merely a source of information about what people actually want a robot to optimize, i.e., the intended reward ~\citep{hadfield2017inverse,ratner2018simplifying}. Luckily, it is not the only one. Robots can also learn about the intended reward from demonstrations (IRL) ~\citep{ng2000algorithms,abbeel2004apprenticeship}, by asking us to make comparisons between trajectories~\citep{wirth2017survey,dorsa2017active,christiano2017deep}, or by grounding our instructions~\citep{macglashan2015grounding,fu2019language}.

Perhaps even more fortunate is that we seem to \emph{leak} information left and right about the intended reward. For instance, if we push the robot away, this shouldn't just modify the robot's \emph{current} behavior -- it should also inform the robot about our preferences more \emph{generally} ~\citep{jain2015learning,bajcsy2017learning}. If we turn the robot off in a state of panic to prevent it from disaster, this shouldn't just stop the robot right now. It should also inform the robot about the intended reward function so that the robot prevents itself from the same disaster in the future: the robot should infer that whatever it was about to do has a tragically low reward. Even the \emph{current state} of the world ought to inform the robot about our preferences -- it is a direct result of us having been acting in the world according to these preferences \citep{shah2019preferences}! For instance, those shoes didn't magically align themselves at the entrance, someone put effort into arranging them that way, so their state alone should tell the robot something about what we want.

Overall, there is much information out there, some purposefully communicated, other leaked. While existing papers are instructing us how to tap into some of it, one can only imagine that there is much more that is yet untapped. There are probably new yet-to-be-invented ways for people to purposefully provide feedback to robots -- e.g. guiding them on which part of a trajectory was particularly good or bad.  And, there will probably be new realizations about ways in which human behavior already leaks information, beyond the state of the world or turning the robot off. How will robots make sense of all these diverse sources of information?

Our insight is that there is a way to interpret all this information in a single unifying formalism. The critical observation is that human behavior is a \emph{reward-rational implicit choice} -- a choice from an implicit set of options, which is approximately rational for the intended reward. This observation leads to a \emph{recipe} for making sense of human behavior, from language to switching the robot off. The recipe has two ingredients: 1) the set of \emph{options} the person (implicitly) chose from, and 2) a \emph{grounding} function that maps these options to robot behaviors. This is admittedly obvious for traditional feedback. In comparison feedback, for instance, the set of options is just the two robot behaviors presented to the human to compare, and the grounding is identity. In other types of behavior though, it is much less obvious. Take switching the robot off. The set of options is implicit: you can turn it off, or you can do nothing. The formalism says that when you turn it off, it should know that you could have done nothing, but (implicitly) chose not to. That, in turn, should propagate to the robot's reward function. For this to happen, the robot needs to ground these options to robot behaviors: identity is no longer enough, because it cannot directly evaluate the reward of an utterance or of getting turned off, but it can evaluate the reward of robot actions or trajectories. Turning the robot off corresponds to a trajectory -- whatever the robot did until the off-button was pushed, followed by doing nothing for the rest of the time horizon. Doing nothing corresponds to the trajectory the robot was going to execute. Now, the robot knows you prefer the former to the latter. We have taken a high-level human behavior, and turned it into a direct comparison on robot trajectories with respect to the intended reward, thereby gaining reward information.

We use this perspective to survey prior work on reward learning. We show that despite their diversity, many sources of information about rewards proposed thus far can be characterized as instantiating this formalism (some very directly, others with some modifications). This offers a unifying lens for the area of reward learning, helping better understand and contrast prior methods. We end with discussion on how the formalism can help combine and actively decide among feedback types, and also how it can be a potentially helpful recipe for interpreting new types of feedback or sources of leaked information. %

%% file: unifying_framework.tex
\section{A formalism for reward learning}

\subsection{Reward-rational implicit choice} \label{sec:formalism}
In reward learning, the robot's goal is to learn a reward function $\reward : \trajs \rightarrow \R$ from human behavior that maps trajectories\footnote{We consider finite fixed horizon T trajectories.} $\traj\in\trajs$ to scalar rewards.%

\textbf{(Implicit/explicit) set of options $\choices$.} We interpret human behavior as choosing an option $\fdbkch$ from a set of options $\choices$. Different behavior types will correspond to different explicit or implicit sets $\choices$. For example, when a person is asked for a \emph{trajectory comparison}, they are explicitly shown two trajectories and they pick one. However, when the person gives a \emph{demonstration}, we think of the possible options $\choices$ as implicitly being all possible trajectories the person could have demonstrated. The implicit/explicit distinction brings out a general tradeoff in reward learning. The cleverness of implicit choice sets is that even when we cannot enumerate and show all options to the human, e.g. in demonstrations, we still rely on the human to optimize over the set. On the other hand, an implicit set is also risky -- since it is not explicitly observed, we may get it wrong, potentially resulting in worse reward inference.

\textbf{The grounding function $\map$.} We link the human's choice to the reward by thinking of the choice as (approximately) maximizing the reward. However, it is not immediately clear what it means for the human to maximize reward when choosing feedback because the feedback may not be a (robot) trajectory, and the reward is only defined over trajectories. For example, in \emph{language feedback}, the human describes what they want in words. What is the reward of the sentence, ``Do not go over the water''?

To overcome this syntax mismatch, we map options in $\choices$ to (distributions over) trajectories with a grounding function $\map : \choices \rightarrow f_\trajs$ where $f_\trajs$ is the set of distributions over trajectories for the robot $\trajs$. Different types of feedback will correspond to different groundings. In some instances, such as kinesthetic demonstrations or trajectory comparisons, the mapping is simply the identity. In others, like corrections, language, or proxy rewards, the grounding is more complex (see Section \ref{sec:applying-formalism}).

\textbf{Human policy.} Given the set of choices $\choices$ and the grounding function $\map$, the human's approximately rational choice $\fdbkch \in \choices$ can now be modeled via a \emph{Boltzmann-rational} policy, a policy in which the probability of choosing an option is exponentially higher based on its reward:
\begin{align} \label{eq:rat-choice2}
   \pr (\fdbkch \mid \reward,\, \choices)=\frac{\exp(\beta \cdot \ex_{\traj \sim \map(\fdbkch)}[\reward(\traj)] )}{\sum_{\fdbk\in \choices}\exp(\beta \cdot \ex_{\traj \sim \map(\fdbk)}[\reward(\traj)])} \,,
\end{align}
where the parameter $\rat$ is a coefficient that models how rational the human is. Often, we simplify Equation \ref{eq:rat-choice2} to the case where $\map$ is a deterministic mapping from choices in $\choices$ to trajectories in $\trajs$, instead of distributions over trajectories. Then, the probability of choosing $\fdbkch$ can be written as:\footnote{One can also consider a variant of Equation \ref{eq:rat-choice2} in which choices are grounded to actions, rather than trajectories, and are evaluated via a Q-value function, rather than the reward function. That is, $\map:\choices \rightarrow \mathcal{A}$, $\pr (\fdbkch \mid \reward,\, \choices)\propto \exp(\beta \cdot \ex_{a \sim \map(\fdbkch)}[Q^*(\st, a)])$.} %
\begin{equation} \label{eq:rat-choice-det}
   \pr (\fdbkch \mid \reward,\, \choices) \propto \exp(\beta \cdot \reward(\map(\fdbkch)))  
\end{equation}

Boltzmann-rational policies are widespread in psychology \citep{baker2009action,goodman2009cause,goodman2013knowledge}, economics \cite{bradley1952rank,luce1959individual,plackett1975analysis,luce1959individual}, and AI \cite{ziebart2008maximum,ramachandran2007bayesian,finn2016guided,bloem2014infinite,dragan2013legibility} as models of human choices, actions, or inferences. But why are they a reasonable model?

While there are many possible motivations, we contribute a derivation (Appendix \ref{app:max-ent}) as the maximum-entropy distribution over choices for a \emph{satisficing} agent, i.e. an agent that in expectation makes a choice with $\epsilon$-optimal reward. A higher value of $\epsilon$ results in a lower value of $\rat$, modeling less optimal humans.

\begin{definition}[Reward-rational choice]
Finally, putting it all together, we call a type of feedback a \emph{reward-rational choice} if, given a grounding function $\map$, it can be modeled as a choice from an (explicit or implicit set) $\choices$ that (approximately) maximizes reward, i.e., as in Equation \ref{eq:rat-choice2}.
\end{definition}

\subsection{Robot inference} \label{sec:robot-inf}
Each feedback is an observation about the reward, which means the robot can run Bayesian inference to update its belief over the rewards. For a determinstic grounding,
\begin{align}
    \pr(\reward \mid \fdbkch) = \frac{1}{Z} \cdot  \frac{\exp(\beta \cdot \reward(\map(\fdbkch)))}{\sum_{\fdbk\in \choices}\exp(\beta \cdot \reward(\map(\fdbk)))} \cdot \pr(\reward) \,,
\label{eq:bayesianupdate}
\end{align}
where $\pr(\reward)$ is the prior over rewards and $Z$ is the normalization over possible reward functions. The inference above is often intractable, and so reward learning work leverages approximations~\citep{blei2017variational}, or computes only the MLE for a parametrization of rewards (more recently as weights in a neural network on raw input ~\citep{christiano2017deep,ibarz2018reward}).

Finally, when the human is highly rational ($\beta \rightarrow \infty$), the only choices in $\choices$ with a non-neglible probability of being picked are the choices that exactly maximize reward. Thus, the human's choice $\fdbkch$ can be interpreted as \emph{constraints} on the reward function (e.g. \citep{ratliff2006maximum}):
\begin{gather} \label{eq:constraints}
    \text{Find } \reward \text{ such that }
    \reward(\map(\fdbkch)) \geq \reward(\map(\fdbk)) \quad \forall \fdbk \in \choices \,.
\end{gather}

\begin{table*}[t!]
\centering
    \captionsetup{justification=centering}
    \caption{The choice set $\choices$ and grounding function $\map$ for different types of feedback described in Section \ref{sec:applying-formalism}, unless otherwise noted.}
    \label{tab:c-grding}
    
    {\renewcommand{\arraystretch}{1.5}
    \resizebox{1\columnwidth}{!}{
    \begin{tabular}{lll}
        \toprule
        \textbf{Feedback} & \textbf{Choices $\choices$} & \textbf{Grounding $\map$} \\
         Comparisons \citep{wirth2017survey} & $\traj_i\in \{\traj_1, \traj_2\}$ & $\map(\traj_i) = \traj_i$\\
        Demonstrations \cite{ng2000algorithms} & $\traj_d \in \trajs$ & $\map(\traj) = \traj$  \\
        Corrections \citep{bajcsy2017learning} & $\Delta q \in Q - Q$ & $\map(\Delta q) = \traj_R + A^{-1}\Delta q$ \\
        Improvement \citep{jain2015learning} & $\xi \in \{\xi_{\text{improved}},\, \xi_R\}$ & $\map(\xi) = \xi$ \\
        Off \citep{hadfield2017off} & $c \in \{\text{off},-\}$ & $\map(c) = \begin{cases} \traj_R & c = - \\ \traj_R^{0:t}\traj_R^t\dots\traj_R^t & c = \text{off} \end{cases}$ \\
        Language \citep{matuszek2012joint} & $\lambda \in \langset$ & $\map(\lang) = \text{Unif}(G(\lang))$ \\
        Proxy Rewards \citep{hadfield2017inverse} & $\tilde{r} \in \tilde{\rewards}$ & $\map(\tilde{r}) = \pi(\traj \mid \tilde{r})$  \\
        Reward and Punishment \citep{griffith2013policy} & $\fdbk \in \{+1, -1\}$ & $\map(c) = \begin{cases} \traj_R & c = +1 \\ \traj_{\text{expected}} & c = -1 \end{cases}$ \\
        Initial state \citep{shah2019preferences} & $s \in \mathcal{S}$ & $\map(s) = \text{Unif}(\{\traj_{H}^{-T:0} \mid \traj_{H}^{0}=s \})$ \\
        Credit assignment (Discussion) & $\traj \in \{\traj_R^{i:i+k},\, 0 \leq i \leq T \}$ & $\map(\traj) = \traj$ \\
        \bottomrule
    \end{tabular}
    }}
\end{table*}

\begin{table*}
    \captionsetup{justification=centering}
    \caption{The probabilistic model (Equation \ref{eq:rat-choice2}) and the simplification to the constraint-based model (Equation \ref{eq:constraints}).}
    \label{tab:policies}
    \centering
    \begingroup
    \renewcommand{\arraystretch}{1.5}
    \resizebox{\textwidth}{!}{%
    \begin{tabular}{lll}
        \toprule
        \textbf{Feedback} & \textbf{Constraint} & \textbf{Probabilistic}\\
         Comparisons &
         $\reward(\traj_{1}) \ge \reward(\traj_{2})$ &
         $\pr(\traj_1 \mid r,\choices) \ = \ \dfrac{\exp(\beta\cdot r(\traj_{1}))}{\exp(\beta\cdot r(\traj_{1}))+\exp(\beta\cdot r(\traj_{2}))}$\\
        Demonstrations &
        $r(\traj_D) \ge r(\traj) \quad \forall \ \traj \in \trajs$ & $\pr(\traj_D \mid r,\trajs) \ = \ \dfrac{\exp(\beta\cdot r(\traj_D))}{\sum_{\traj\in\trajs}\exp(\beta\cdot r(\traj))}$\\
        Corrections &
        $r(\traj_R + A^{-1}\Delta q) \ge r(\traj_R+ A^{-1}\Delta q')\ \forall \Delta q' \in Q - Q$ &
        $\pr(\Delta q' \mid r, Q-Q)  = \dfrac{\exp(\beta\cdot r(\traj_R+ A^{-1}\Delta q))}{\sum_{\Delta q \in Q-Q} \exp(\beta\cdot r(\traj_R+ A^{-1}\Delta q))}$\\
        Improvement &
        $r(\traj_{\text{improved}}) \ge r(\traj_R)$
        & $\pr(\traj_{\text{improved}} \mid r, \choices) \ = \ \dfrac{\exp(\beta\cdot r(\traj_{\text{improved}}))}{\exp(\beta\cdot r(\traj_{\text{improved}})) + \exp(\beta\cdot r(\traj_{R}))}$\\
        Off &
        $r(\traj_\rob^{0:t}\traj^t\dots\traj^t) \ge r(\traj_R)$ &
        $\pr(\text{off} \mid r, \choices) \ = \ \dfrac{\exp(\beta\cdot r(\traj_\rob^{0:t}\traj^t\dots\traj^t))}{\exp(\beta\cdot r(\traj_\rob^{0:t}\traj^t\dots\traj^t))+\exp(\beta\cdot r(\traj_\rob))}$\\
        Language &
        $\mathbb{E}_{\traj \sim \text{Unif}(G(\lang^*))}\big[r(\traj)\big] \ge \mathbb{E}_{\traj \sim \text{Unif}(G(\lang))}\big[r(\traj)\big]\ \forall \lambda\in\langset$ &
        $\pr(\lambda^* \mid r,\langset) \ = \ \dfrac{\exp(\beta\cdot \mathbb{E}_{\traj \sim \text{Unif}(G(\lang^*))}\big[r(\traj) \big])}{\sum_{\lambda\in\langset}\exp(\beta\cdot \mathbb{E}_{\traj \sim \text{Unif}(G(\lang))}\big[r(\traj) \big])}$\\
        Proxy Rewards &
        $\mathbb{E}_{\tilde{\traj}\sim\pi(\tilde{\traj} \mid \tilde{r})}\big[r(\tilde{\traj})\big] \ge \mathbb{E}_{\tilde{\traj}\sim\pi(\tilde{\traj} \mid c)}\big[r(\tilde{\traj})\big] \quad \forall c \in \tilde{\rewards}$ & 
        $\pr(\tilde{r}\mid \reward, \tilde{\rewards})\ = \ \dfrac{\exp(\beta\cdot\mathbb{E}_{\tilde{\traj}\sim\pi(\tilde{\traj} \mid \tilde{r})}\big[r(\tilde{\traj})\big])}{\sum_{c\in \tilde{\rewards}}\exp(\beta\cdot\mathbb{E}_{\tilde{\traj}\sim\pi(\tilde{\traj} \mid c)}\big[r(\tilde{\traj}) \big])} $
        \\
        Reward/Punish & $\reward(\traj_R) \geq \reward(\traj_{\text{expected}})$ & $\pr(+1 \mid \reward, \choices) = \dfrac{\exp(\beta \cdot \reward(\traj_{R}))}{\exp(\beta \cdot \reward(\traj_R)) + \exp(\beta \cdot \reward(\traj_{\text{expected}}))}$ \\
        Initial state & $\ex_{\traj \sim \map(s^*)}[r(s^*)] \geq \ex_{\traj \sim \map(s)}[r(s)] \quad \forall s \in \mathcal{S}$ & $\pr(s^* \mid r, \mathcal{S}) = \dfrac{\exp(\beta \cdot \ex_{\traj \sim \map(s^*)}[r(\traj)])}{\sum_{s \in S} \exp(\beta \cdot \ex_{\traj \sim \map(s)}[r(\traj)])}$ \\ 
        Meta-choice & $\ex_{\traj \sim \map(\choices_i)}[\reward(\traj)] \geq \ex_{\traj \sim \map(\choices_j)}[\reward(\traj)] \quad \forall j \in [n]$ & $\pr(\choices_i \mid r, \choices_0) = \dfrac{\exp\big( \beta_0\cdot\mathbb{E}_{\traj \sim \map_0(\choices_i)}[r(\traj)] \big)}{\sum_{j \in [n]}\ \exp\big( \beta_0\cdot\mathbb{E}_{\traj \sim \map_0(\choices_j)}[r(\traj)] \big)}$\\
        Credit assignment &  $r(\traj^*) \ge r(\traj) \quad \forall \ \traj \in \choices$ & $\pr(\traj^* \mid r,\choices) \ = \ \dfrac{\exp(\beta\cdot r(\traj^*))}{\sum_{\traj\in\choices}\exp(\beta\cdot r(\traj))}$ \\
        \bottomrule
    \end{tabular}}
    \endgroup
\end{table*}

\begin{figure*}[t]
    \centering
    \includegraphics[width=\textwidth]{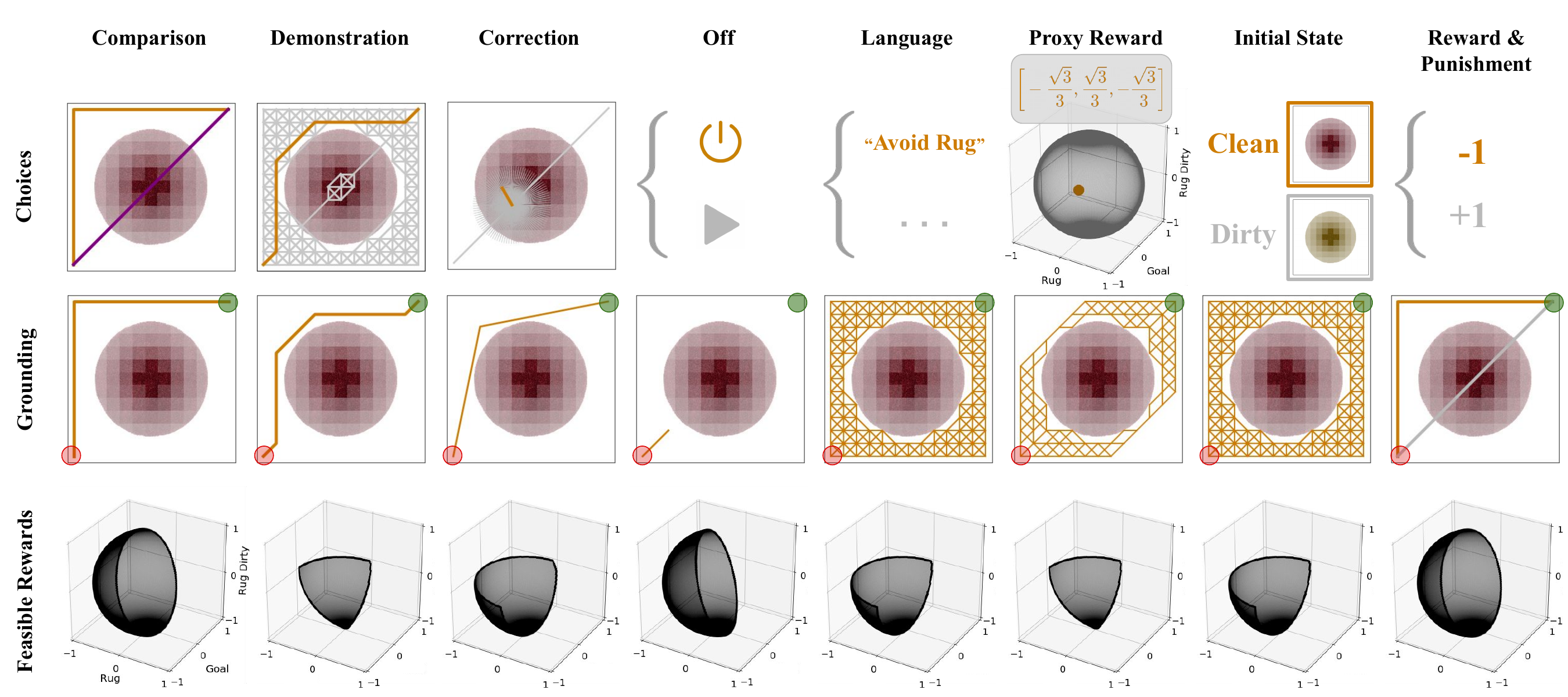}
    \caption{\small{Different behavior types described in \sref{sec:applying-formalism} in a gridworld with three features: avoiding/going on the rug, getting the rug dirty, and reaching the goal (green). For each, we display the choices, grounding, and feasible rewards under the constraint formulation of robot inference (\ref{eq:constraints}). 
    Each trajectory is a finite horizon path that begins at the start (red). Orange is used to denote $\fdbkch$ and $\map(\fdbkch)$ while gray to denote other choices $c$ in $\choices$.} For instance, the \textbf{comparison} affects the feasible reward space by removing the halfspace where going on the rug is good. It does not inform the robot about the goal, because both end at the goal. The \textbf{demonstration} removes the space where the rug is good, where the goal is bad (because alternates do not reach the goal), and where getting the rug dirty is good (because alternates slightly graze the rug). The \textbf{correction} is similar to the demonstration, but does not infer about the goal, since all corrections end at goal.}
    \label{fig:TableFig}
\end{figure*}

%% file: applying_formalism.tex
\section{Prior work from the perspective of the formalism} \label{sec:applying-formalism}
We now instantiate the formalism above with different behavior types from prior work, constructing their choice sets $\choices$ and groundings $\map$. Some are obvious -- comparisons, demonstrations especially. Others -- initial state, off, reward/punish -- are more subtle and it takes slightly modifying their original methods to achieve unification, speaking to the nontrivial nuances of identifying a common formalism.%

Table \ref{tab:c-grding} lists $\choices$ and $\map$ for each feedback, while Table \ref{tab:policies} shows the deterministic constraint on rewards each behavior imposes, along with the probabilistic observation model -- highlighting, despite the differences in feedback, the pattern of the (exponentiated) choice reward in the numerator, and the normalization over $\choices$ in the denominator. Fig. \ref{fig:TableFig} will serve as the illustration for these types, looking at a grid world navigation task around a rug. The space of rewards we use for illustration is three-dimensional weight vectors for avoiding the rug, not getting dirty, and reaching the goal.

\textbf{Trajectory comparisons.} In trajectory comparisons \citep{wirth2016model}, the human is typically shown two trajectories $\traj_1 \in \trajs$ and $\traj_2 \in \trajs$, and then asked to select the one that they prefer. They are perhaps the most obvious exemplar of reward-rational choice: the set of choices $\choices = \{\traj_1, \traj_2\}$ is explicit, and the grounding $\map$ is simply the identity. As Fig. \ref{fig:TableFig} shows, for linear reward functions, a comparison corresponds to a hyperplane that cuts the space of feasible reward functions in half. For all the reward functions left, the chosen trajectory has higher reward than the alternative. Most work on comparisons is done in the preference-based RL domain in which the robot might compute a policy directly to agree with the comparisons, rather than explicitly recover the reward function \citep{wilson2012bayesian,busa2013preference}. Within methods that do recover rewards, most use the constraint version (left column of Table \ref{tab:policies}) using various losses \citep{akrour2011preference, wirth2014learning}. \citep{holladay2016active} uses the Boltzmann model (right column of Table \ref{tab:policies}) and proposes actively generating the queries, \citep{dorsa2017active} follows up with actively synthesizing the queries from scratch, and \citep{christiano2017deep} introduces deep neural network reward functions.

\textbf{Demonstrations.} In demonstrations, the human is asked to demonstrate the optimal behavior. Reward learning from demonstrations is often called \emph{inverse reinforcement learning} (IRL) and is one of the most established types of feedback for reward learning \citep{ng2000algorithms,abbeel2004apprenticeship,ziebart2008maximum}. 
Unlike in comparisons, in demonstrations, the human is not explicitly given a set of choices. However, we assume that the human is \emph{implicitly} optimizing over all possible trajectories (Fig. \ref{fig:TableFig} (1st row, 2nd column) shows these choices in gray). Thus, demonstrations are a reward-rational choice in which the set of choices $\choices$ is (implicitly) the set of trajectories $\trajs$. Again, the grounding $\map$ is the identity. In \figref{fig:TableFig}, fewer rewards are consistent with a demonstration than with a comparison.
Early work used the constraint formulation with various losses to penalize violations \citep{ng2000algorithms,ratliff2006maximum}. Bayesian IRL \citep{ramachandran2007bayesian} exactly instantiates the formalism using the Boltzmann distribution by doing a full belief update as in Equation \ref{eq:bayesianupdate}. Later work computes the MLE instead \citep{ziebart2008maximum, bloem2014infinite, ho2016generative} and approximates the partition function (the denominator) by a quadratic approximation about the demonstration \citep{levine2012continuous}, a Laplace approximation \citep{dragan2012formalizing}, or importance sampling \citep{finn2016guided}.

\textbf{Corrections} are the first type of feedback we consider that has both an implicit set of choices $\choices$ and a non-trivial (not equal to identity) grounding. Corrections are most common in physical human-robot interaction (pHRI), in which a human physically corrects the motion of a robot. The robot executes a trajectory $\traj_\rob$, and the human intervenes by applying a correction $\Delta q \in Q$ that modifies the robot's current configuration. Therefore, the set of choices $\choices = Q - Q$ consists of all possible configuration differences $\Delta q$ the person could have used (\figref{fig:TableFig} 1st row, 3rd column shows possible $\Delta q$s in gray and the selected on in orange). The way we can ground these choices is by finding a trajectory that is closest to the original, but satisfies the constraint of matching a new point:
\begin{align}
    & \min_{\traj} ||\traj-\traj_\rob||^2_{A} \nonumber\\
    & s.t. \ \traj(0)=\traj_\rob(0),\ \traj(T)=\traj_\rob(T), \traj(t)=\traj_\rob(t)+\Delta q
\end{align}
where $t$ is the time at which the correction was applied. Choosing a non-Euclidean inner-product, A (for instance $K^TK$, with $K$ the finite differencing matrix), couples states along the trajectory in time and leads to a the resulting trajectory smoothly deforming -- propagating the change $\Delta q$ to the rest of the trajectory:  $\map(\Delta q)=\traj_\rob+A^{-1}[\lambda,0,.., \Delta q,..,0,\gamma]^T$ (with $\lambda$ and $\gamma$ making sure the end-points stay in place).
This is the orange trajectory in the figure. 
Most work in corrections affects the robot's trajectory but not the reward function \citep{haddadin2008collision,hogan1985impedance}, with \citep{losey2017trajectory} proposing the propagation via $A^{-1}$ above. \cite{bajcsy2017learning} propose that corrections are informative about the reward and use the propagation as their grounding, deriving an approximate MAP estimate for the reward. \cite{losey2018including} introduce a way to maintain uncertainty.

\textbf{Improvement.} Prior work \citep{jain2015learning} has also modeled a variant of corrections in which the human provides an improved trajectory $\traj_{improved}$ which is treated as better than the robot's original $\traj_\rob$. Although \citep{jain2015learning} use the Euclidean inner product and implement reward learning as an online gradient method that treats the improved trajectory as a demonstration (but only takes a single gradient step towards the MLE), we can also naturally interpret improvement as a comparison that tells us the improved trajectory is better than the original: the set of options $\choices$ consists of only $\traj_\rob$ and $\traj_{improved}$ now, as opposed to all the trajectories obtainable by propagating local corrections; the grounding is identity, resulting in essentially a comparison between the robot's trajectory and the user provided one.

\textbf{Off.} In ``off'' feedback, the robot executes a trajectory, and at any point, the human may switch the robot off. ``Off'' appears to be a very sparse signal, and it is not spelled out in prior work how one might learn a reward from it. Reward-rational choice suggests that we first uncover the implicit set of options $\choices$ the human was choosing from. In this case, the set of options consists of turning the robot off or not doing anything at all: $\choices = \{\text{off}, -\}$. Next, we must ask how to evaluate the reward of the two options, i.e., what is the grounding? \cite{hadfield2017off} introduced off feedback and formalized it as a choice for a one-shot game. There, not intervening means the robot takes its one possible action, and intervening means the robot takes the no-op action. This can be easily generalized to the sequential setting: not intervening means that the robot continues on its current trajectory, and intervening means that it stays at its current position for the remainder of the time horizon. Thus, the choices $\choices =\{\text{off}, -\}$ map to the trajectories $\{\xi_\rob^{0:t}\traj^{t}_\rob\dots \traj^{t}_\rob,\, \traj_\rob \}$.

\textbf{Language.} Humans might use rich language to instruct the robot, like ``Avoid the rug.'' Let $G(\lang)$ be the trajectories that are consistent with an utterance $\lang \in \langset$ (e.g. all trajectories that do not enter the rug). 
Usually the human instruction is interpreted \emph{literally}, i.e. any trajectory consistent with the instruction $\traj \in \lgrding(\lang)$ is taken to be equally likely , although, other distributions are also possible. For example, a problem with literal interpretation is that it does not take into account the other choices the human may have considered. The instruction ``Do not go into the water'' is consistent with the robot not moving at all, but we imagine that if the human wanted the robot to do nothing, they would have said that instead. Therefore, it would be incorrect for the robot to do nothing when given the instruction ``Do not go into the water''. This type of reasoning is called \emph{pragmatic reasoning}, and indeed recent work shows that explicitly interpreting instructions pragmatically can lead to higher performance \citep{fried2018unified,fried2018speaker}. The reward-rational choice formulation of language feedback naturally leads to pragmatic reasoning on the part of the robot, and is in fact equivalent to the rational speech acts model \citep{goodman2013knowledge}, a standard model of pragmatic reasoning in language. The pragmatic reasoning arises because the human is explicitly modeled as choosing from a set of options.

The reward-rational choice formulation of language feedback naturally leads to pragmatic reasoning on the part of the robot, and is in fact equivalent to the rational speech acts model \citep{goodman2013knowledge}, a standard model of pragmatic reasoning in language. The pragmatic reasoning arises because the human is explicitly modeled as choosing from a set of options.
Language is a reward-rational choice in which the set of options $\choices$ is the set of instructions considered in-domain $\langset$ and the grounding $\map$ maps an utterance $\lang$ to the uniform distribution over consistent trajectories $\text{Unif}(G(\lang))$. In language feedback, a key difficulty is learning which robot trajectories are consistent with a natural language instruction, the \emph{language grounding problem} (and is where we borrow the term ``grounding'' from) \citep{matuszek2012joint,tellex2011understanding,fu2019language}. \figref{fig:TableFig} shows the grounding for avoiding the rug in orange -- all trajectories from start to goal that do not enter rug cells. %

\textbf{Proxy rewards} are expert-specified rewards that do not necessarily lead to the desired behavior in all situations, but can be trusted on the training environments. They were introduced by \cite{hadfield2017inverse}, who argued that even when the expert attempts to fully specify the reward, it will still fail to generalize to some situations outside of the training environments.  Therefore, rather than taking a specified reward at face value, we can interpret it as evidence about the true reward. Proxy reward feedback is a reward-rational choice in which the set of choices $\choices$ is the set of proxy rewards the designer may have chosen, $\tilde{\rewards}$. The reward designer is assumed to be approximately-optimal, i.e. they are more likely to pick a proxy reward $\tilde{\reward} \in \tilde{\rewards}$ if it leads to better trajectories \emph{on the training environment(s)}. Thus, the grounding $\map$ maps a proxy reward $\reward$ to the distribution over trajectories that the robot takes in the training environment given the proxy reward \citep{hadfield2017inverse,activeird:2018,ratner2018simplifying}. \figref{fig:TableFig} shows the grounding for a proxy reward for reaching the goal, avoiding the rug, and not getting the rug dirty -- many feasible rewards would produce similar behavior as the proxy. By taking the proxy as evidence about the underlying reward, the robot ends up with uncertainty over what the actual reward might be, and can better hedge its bets at test time. 

\textbf{Reward and punishment} \citep{griffith2013policy,loftin2014strategy}. In this type of feedback, the human can either reward ($+1$) or punish ($-1$) the robot for its trajectory $\traj_R$; the set of options is $\choices = \{+1, -1\}$. A naive implementation would interpret reward and punishment literally, i.e. as a scalar reward signal for a reinforcement learning agent, however empirical studies show that humans reward and punish based on how well the robot performs \emph{relative to their expectations} \citep{macglashan2017interactive}. Thus, we can use our formalism to interpret that: reward ($+1$) grounds to the robot's trajectory $\traj_R$, while punish ($-1$) grounds to the trajectory the human expected $\traj_{\text{expected}}$ (not necessarily observed).

\textbf{Initial state.}  \citet{shah2019preferences} make the observation that when the robot is deployed in an environment that humans have acted in, the current state of the environment is already optimized for what humans want, and thus contains information about the reward. For example, suppose the environment has a goal state which the robot can reach through either a paved path or a carpet. If the carpet is pristine and untrodden, then humans must have intentionally avoided walking on it in the past (even though the robot hasn't observed this past behavior), and the robot can reasonably infer that it too should not go on the carpet. 

The original paper inferred rewards from a single state $s$ by marginalizing over possible pasts, i.e. trajectories $\traj_{H}^{-T:0}$ that end at $s$ which the human could have taken, $P(s|r)=\sum_{\traj_{H}^{-T:0}|\traj_{H}(0)=s}P(\traj_{H}^{-T:0}|r)$. However, through the lens of our formalism, we see that initial states can also be interpreted more directly as reward-rational implicit choices. The set of choices $\choices$ can be the set of possible initial states $\mathcal{S}$. The grounding function $\map$ maps a state $s \in \mathcal{S}$ to the uniform distribution over any human trajectory $\traj_{H}^{-T:0}$ that starts from a specified time before the robot was deployed ($t=-T$) and ends at state $s$ at the time the robot was deployed ($t=0$), i.e. $\traj_H^{0}=s$. This leads to the $P(s|r)$ from Table \ref{tab:policies}, which is almost the same as the original, but sums over trajectories directly in the exponent, and normalizes over possible other states. The two interpretations would only become equivalent if we replaced the Boltzmann distribution with a linear one.
 Fig \ref{fig:TableFig} shows the result of this (modified) inference, recovering as much information as with the correction or language.

%% file: disc.tex
\section{Discussion of implications} \label{sec:disc}
From demonstrations to reward/punishment to the initial state of the world, the robot can extract information from humans by modeling them as making approximate reward-rational choices. Often, the choices are implicit, like in turning the robot off or providing language instructions. Sometimes, the choices are not made in order to purposefully communicate about the reward, and rather end up leaking information about it, like in the initial state, or even in corrections or turning the robot off.  Regardless, this unifying lens enables us to better understand, as in \figref{fig:TableFig}, how all these sources of information relate and compare.

Down the line, we hope this formalism will enable research on combining and actively querying for feedback types, as well as making it easier to do reward learning from new, yet to be uncovered sources of information. Concretely, so far we have talked about learning from individual types of behaviors. But we do not want our robots stuck with a single type: we want them to 1) read into all the leaked information, and 2) learn from all the purposeful feedback. For example, the robot might receive demonstrations from a human during training, and then corrections during deployment, which were followed by the human prematurely switching the robot off. The observational model in (\ref{eq:rat-choice-det}) for a single type of behavior also provides a natural way to model combinations of behavior. If each observation is conditionally independent given the reward, then according to (\ref{eq:rat-choice-det}), the probability of observing a vector $\mathbf{\fdbk}$ of $n$ behavioral signals (of possibly different types) is equal to
\begin{align} \label{eq:prob-combine}
    \pr(\mathbf{\fdbk} \mid r) = \prod_{i=1}^{n} \frac{\exp(\rat_i \cdot \reward (\map_i(\mathbf{\fdbk}_i)))}{\sum_{\fdbk \in \choices_i} \exp(\rat_i \cdot \reward (\map_i(\fdbk)))} \,.
\end{align}
Given this likelihood function for the human's behavior, the robot can infer the reward function using the approaches and approximations described in \sref{sec:robot-inf}. Recent work has already built in this direction, combining trajectory comparisons and demonstrations \citep{ibarz2018reward,palan2019learning}. We note that the formulation in Equation \ref{eq:prob-combine} is general and applies to \emph{any} combination. In Appendix \ref{app:case-study}, we describe a case study on a novel combination of feedback types: proxy rewards, a physical improvement, and comparisons in which we use a constraint-based approximation (see Equation \ref{eq:constraints}) to Equation \ref{eq:prob-combine}.

Further, it also becomes natural to \emph{actively decide} which feedback type to ask a human for. Rather than relying on a heuristic (or on the human to decide), the robot can maximize expected information gain.  Suppose we can select between $n$ types of feedback with choice sets $\choices_1, \dots, \choices_n$ to ask the user for. Let $b_t$ be the robot's belief distribution over rewards at time $t$. The type of feedback $i^*$ that (greedily) maximizes information gain for the next time step is
\begin{equation} 
    i^* = \argmax_{i \in [n]}  \ex_{\reward_t, \fdbkch_i} \left [\log \left (\frac{p(\fdbkch_i \mid \reward_t)}{\int_{\reward_t \in \rewards}p(\fdbkch_i \mid \reward_t)b_t(\reward_t)} \right) \right]\,,
   \label{eq:info}
\end{equation}
where $\reward_t \sim B_t$ is distributed according to the robot's current belief, $\fdbkch_i \in \choices_i$ is the random variable corresponding to the user's choice within feedback type $i$, and $p(\fdbkch_i \mid \reward_t)$ is defined according to the human model in Equation \ref{eq:rat-choice2}. We also note that different feedback types may have different costs associated with them (e.g. of human time) and it is straight-forward to integrate these costs into (\ref{eq:info}). In Appendix \ref{app:active}, we describe experiments with active selection of feedback types. In the environments we tested, we found that demonstrations are optimal early on, when little is known about the reward, while comparisons became optimal later, as a way to fine-tune the reward. The finding provides validation for the approach pursued by \cite{palan2019learning} and \cite{ibarz2018reward}. Both papers manually define the mixing procedure we found to be optimal: initially train the reward model using human demonstrations, and then fine-tune with comparisons. 

Finally, the types of feedback or behavior we have discussed so far are by no means the only types possible. New ones will inevitably be invented. But when designing a new type of feedback, it is often difficult to understand what the relationship is between the reward $\reward$ and the feedback $\fdbkch$. Reward-rational choice suggests a recipe for uncovering this link -- define what the implicit set of options the human is choosing from is, and how those options ground to trajectories. Then, Equation \ref{eq:rat-choice2} provides a formal model for the human feedback.

For example, hypothetically, someone might propose a ``credit assignment" type of feedback. Given a trajectory $\traj_R$ of length $T$, the human is asked to pick a segment of length $k < T$ that has maximal reward. We doubt the set of choices in an implementation of credit assignment would be explicit, however the implicit set of choices $\choices$ is then the set of all segments of length $k$. The grounding function $\map$ is simply the identity. With this choice of $\choices$ and $\map$ in hand, the human can now be modeled according to Equation \ref{eq:rat-choice2}, as we show in the last rows of Tables \ref{tab:c-grding} and \ref{tab:policies}.

While of course the formalism won't apply to \emph{all} types of feedback, we believe that it applies to \emph{many}, even to types that initially seem to have a more obvious, literal interpretation (e.g. reward and punishment, Section \ref{sec:applying-formalism}). Most immediately, we are excited about using it to formalize a particular new source of (leaked) information we uncovered while developing the formalism itself: the moment we enable robots to learn from multiple types of feedback, users will have the \emph{choice} of which feedback to provide. Interpreted literally, each feedback gives the robot evidence about the reward. However, this leaves information on the table: if the person decided to, say, turn the robot off, they \emph{implicitly} decided to \emph{not} provide a correction, or use language. Intuitively, this means that turning off the robot was a more appropriate intervention with respect to the true reward.  Interpreting the feedback type itself as reward-rational implicit choice has the potential to enable robots to extract more information about the reward from the same data. We call the choice of feedback type ``meta-choice''. In Appendix \ref{app:meta}, we formalize meta-choice and conduct experiments that showcase its potential importance.

Overall, we see this formalism as providing conceptual clarity for existing and future methods for learning from human behavior, and a fruitful base for future work on multi-behavior-type reward learning.

%% file: appendix.tex
\begin{appendices}

\section{Bounded rationality, maximum entropy, and Boltzmann-rational policies} \label{app:max-ent}
\input{boltzmann.tex}

\input{case_study}

\input{active}

\input{meta}

\end{appendices}

%% file: boltzmann.tex
A perspective on reward learning that makes use at its core the Boltzmann model from Equation \ref{eq:rat-choice2} would not be complete without a formal justification for it within our context. In this section, we derive it as the maximum-entropy distribution for the choices made by a bounded, \emph{satisficing} human.  Our explanation is complementary to that of \citep{ortega2013thermodynamics} who derive an axiomatic, thermodynamic framework to modeling bounded-rational decision making. Their framework leads to much the same interpretation of the Boltzmann-rational distribution, but is significantly more complex than needed for our purposes.

A perfectly rational human choosing from the set $\choices$ would always pick the choice with optimal reward, $\max_{\fdbk \in \choices} \reward(\map(\fdbk))$. However, since humans are bounded, we do not expect them to perform optimally. Herbert Simon proposed the influential idea that humans are \emph{bounded rational} and merely \emph{satisfice} \citep{simon1956rational}, rather than maximize, i.e., they pick an option above some satisfactory threshold, rather than picking the best possible option.

We can abstractly model a satisficing human by modeling their expected reward as equal to a satisficing threshold, $\max_{\fdbk \in \choices} \reward(\map(\fdbk)) - \epsilon$ where $\epsilon \in (0, \epsilon_{\max})$ is the amount of expected error. The maximum possible error, $\epsilon_{max} = \min_{\fdbk \in \choices} \reward(\map(\fdbk)) - \max_{\fdbk \in \choices} \reward(\map(\fdbk))$, corresponds to \emph{anti-rationality}, i.e., always picking the worst option.

Given the constraint that the human's expected reward is satisfactory, how should we pick a distribution to model the human's choices? The principle of maximum entropy \citep{jaynes1957information} gives us a guide. If we want to encode no extra information in the distribution, then we ought to pick the distribution that maximizes entropy subject to the constraint on the satisficing threshold. 

\begin{definition}[Satisficing MaxEnt problem] \label{eq:max-ent} Let be be a distribution $P$ over choice set $\choices$ and let $p$ be a density for $P$ with respect to a base measure $F$. The Shannon entropy of $P$ is defined as $H(P) = -\int_{\choices} p(f)\log p(f) dF(f)$. The \emph{satisficing maximum entropy problem} is to find a distribution $P$ that maximizes entropy subject to the satisficing constraint (\ref{eq:sat-cond}):
\begin{gather}
    \max_P H(P) \nonumber \\
    \text{subject to } \nonumber \\
    \ex_{\fdbk \sim P}[\reward(\map(\fdbk))] = \max_{\fdbk \in \choices}\reward(\map(\fdbk)) - \epsilon \,. \label{eq:sat-cond}
\end{gather}
\end{definition}

It is well-known that the maximum-entropy distribution subject to linear constraints (such as a constraint on the mean like in (\ref{eq:sat-cond})) is the unique exponential distribution that satisfies the constraints. Thus, for our special case, the maximum-entropy distrbution is the Boltzmann distribution with rationality coefficient $\rat$ satisfying the satisficing constraint.
\begin{theorem}[\citep{jaynes1957information}]
The solution to the satisficing maximum entropy problem is the Boltzmann distribution $\pr_\rat(f) \propto \exp (\rat \cdot \reward(\map(\fdbk)))$ where $\rat$ is the unique value satisfying the satisficing constraint (\ref{eq:sat-cond}).
\end{theorem}

Since the expected reward $\ex_{\rat}[\reward(\map(\fdbk)]$ is monotonically increasing in the rationality parameter $\rat$, the satisficing error $\error$ and rationality coefficient $\rat$ have a one-to-one relationship, as summarized in the following corollary.
\begin{cor}
The solution to the satisficing maximum entropy problem is a Boltzmann-rational policy where the rationality coefficient $\rat$ is monotonically decreasing in the satisficing error $\epsilon$. In particular, we have the following:
\begin{center}
\begin{tabular}{lll}
\toprule
     \textbf{Human type} & \textbf{Error $\epsilon$} & \textbf{Rationality  $\rat$} \\ \hline
     Perfectly rational & $\epsilon \rightarrow 0$ & $\rat \rightarrow +\infty$ \\ \hline 
     Random & $\begin{array} {ll} \epsilon = & \max_{\fdbk \in \choices} \reward(\map(\fdbk)) ~- \\ &  \ex_{\fdbk \sim \text{Unif}(\choices)}[\reward(\map(\fdbk))] \end{array}$ & $\beta = 0$ \\ \hline 
     Anti-rational & $\epsilon \rightarrow \epsilon_{\text{max}}$ & $\beta \rightarrow -\infty$ \\
\bottomrule
\end{tabular}
\end{center}
\end{cor}

Thus, we see that the idea of bounded rationality, as in satisficing, and Boltzmann-rationality are in fact equivalent. By following the principle of maximum entropy, Boltzmann-rationality provides a way to model a satisficing human, without implicitly adding in any other assumptions about the human's choice.

%% file: case_study.tex
\section{A case study on combining feedback types} \label{app:case-study}

\begin{figure*}
    \centering
    \includegraphics[width=\textwidth]{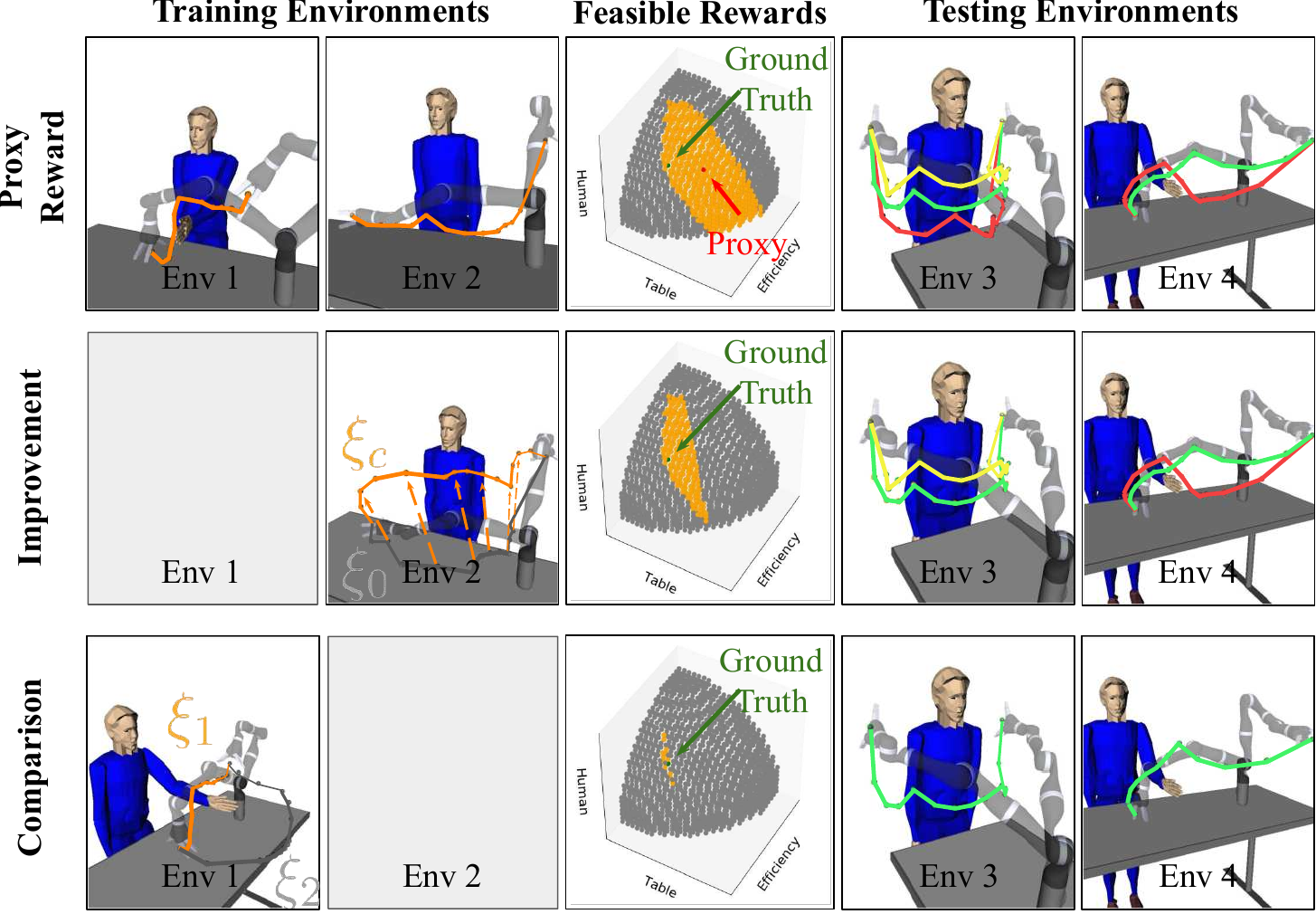}\\
    \caption{\small{A case study for teaching a reward for robot arm motion using two training environments. The robot trades off efficiency, keeping distance away from the human, and also from the table. We use the constraints interpretation of  feedback in this study. We start by defining a proxy reward that produces acceptable behavior (orange trajectories) in the training environments (1st row). This initial feedback significantly prunes the feasible space, but is not enough to guarantee good performance in other environments. On the right, we see trajectories still considered feasible in two test environments. The green one is correct, however, the other feasible trajectories are either too close to the human or too close to the robot. After an improvement feedback and a comparison, the robot shrinks the space of feasible rewards, removing extraneous rewards that produce undesirable behavior at test time.}}
    
    \label{fig:case_study}
\end{figure*}

\figref{fig:case_study} illustrates a case study for teaching a robot arm a reward for motion planning through a novel combination of feedback types. In each environment, the robot arm must plan a trajectory from a start configuration to a designated goal configuration. We want this trajectory to properly trade off efficiency against staying at an appropriate distances to the human, and to the table. Hand-tuning a reward function that returns desirable trajectories in \emph{all} possible environments is actually very challenging. You could imagine that as you increase the efficiency weight to produce a smoother trajectory in one environment, you break the behavior in another environment where the robot now gets too close to the human, etc.  In fact, the first type of feedback in the case study illustrates this: we design a (proxy) reward function that works well in two (training) environments (top left), but there are many rewards that are consistent with that behavior, yet produce vastly different behaviors in the two test environments (right). %

Therefore, we start by defining a proxy reward, but then follow it up with more feedback: an improvement, and a comparison between two trajectories. This narrows down the space of rewards such that the robot can now generalize what to do outside of the training environments, as shown by two testing environments (right). 

\textbf{Cost Function and Features.}
\begin{align*}
    \text{Efficiency} := \sum_{i=1}^{|\tau|-1} \ \|\tau[i]-\tau[i-1]\|^2_2 \,,
\end{align*}
\begin{gather*}
    \text{Distance to Table} := \sum_{i=0}^{|\tau|-1} \ 1-\exp(-\text{dist}_i) \\
    \text{where } \text{dist}_i = |g_{st}(\tau[i])_z - \text{table}_z| \,,
\end{gather*}
\begin{gather*}
    \text{Distance to Human} := \sum_{i=0}^{|\tau|-1} \ 1-\exp(-\text{dist}_i)\\ \text{where } \text{dist}_i = \| \text{proj}_x(g_{st}(\tau[i])) - \text{proj}_x(\text{human}) \|^2_2 \,.
\end{gather*}
Efficiency is the sum of squared configuration space distances between consecutive trajectory waypoints. The table and human features are expressed as 1 minus a radial basis function of a modified distance between the object and the robot's end effector positon (denoted by $g_{st}(\tau[i])$, where $g_{st}$ is the forward kinematics that maps configuration $\traj[i] \in Q$ to its end effector location in $\mathbb{R}^3$). For the table, this modification is to only consider distance in the z-coordinate, effectively measuring the distance from the robot's end effector to the table \emph{plane}. For the human, the modification is to treat the human as an axis $x$ and consider distance in 2 dimensions after projecting onto the plane with normal $x$. In Figure \ref{fig:case_study}, the main obstacle is either the human's body or his arm. When the body is the obstacle, $x=[0,0,1]$ and when the arm is the obstacle, $x=[0, 1,0]$. This considers the human not as just a point, but rather a line along the body, or arm axis.\newline

\textbf{Optimization}
We approximate the space of reward parameters $\Theta$ by uniform discretization at the surface of the non-negative octant of the 3 dimensional sphere (1371 points). Robotic motion planners cannot, in general, compute the globally optimal trajectory for a given $\theta\in \Theta$ so we resort to computing a set $\hat{\mathcal{T}}$ of locally optimal trajectories for each $\theta$ via TrajOpt \citep{Schulmanetal_IJRR2014}. The optimal trajectory for a given $\theta$ is then defined as
$$\traj(\theta) := \argmin_{\traj \in \hat{\mathcal{T}}}\ \theta^T\phi(\traj)$$

\textbf{Proxy Reward.}
For this case study, the robot begins by asking the human designer for a proxy reward (cost) function. It is difficult for humans to provide proxies that work across all environments~\citep{ratner2018simplifying}, so the robot asks for a proxy that produces the desired behavior in the two training environments. The human can provide the proxy weights: $[0.55, 0.55, 0.55]$ and produce trajectories that match those of $\traj_{\theta^*}$ (Figure \ref{fig:case_study} depicted in orange). Providing a proxy applies constraints that shrink our feasible set from $\Theta$ to $\mathcal{F}_{\text{proxy}}$:
\begin{align*}
    \mathcal{F}_{\text{proxy}} = \{\tilde{\theta}\ : \ &  {\theta^*}^T\phi(\traj^{(1)}_{\tilde{\theta}}) \ge {\theta^*}^T\phi(\traj^{(1)}_{\theta}) \,,\,  {\theta^*}^T\phi(\traj^{(2)}_{\tilde{\theta}}) \ge {\theta^*}^T\phi(\traj^{(2)}_{\theta}) \quad \forall\ \theta \in \Theta \} \,,
\end{align*}
where $\traj^{(i)}_{\theta}$ denotes the optimal trajectory\footnote{In our case study, the optimal trajectory is unique.} w.r.t. cost parameter $\theta$ in environment $i$. The new feasible set $\mathcal{F}_{\text{proxy}}$ contains only the parameters $\theta$ that produce optimal trajectories with respect to the true weights $\theta^*$ in environments 1 and 2. Although it is a subset of the original feasible set $\Theta$, the new feasible set $\mathcal{F}_{\text{proxy}}$ is still a reasonably large set (Figure \ref{fig:case_study}, top, middle, orange area). Furthermore, although the proxy  produces optimal trajectories in environments 1 and 2, it does not necessarily for environments 3 and 4. Figure \ref{fig:case_study} (top, right) illustrates the different trajectories that result from optimizing different $\theta\in\mathcal{F}_{\text{proxy}}$. To further narrow our feasible set, we will ask for another form of feedback: Improvement.

\textbf{Improvement.}
The robot will now (actively) provide a nominal trajectory, and ask the human to improve it, i.e. alter the trajectory to better suit their preferences. Suppose the robot presents the human with the nominal trajectory shown in gray (Figure \ref{fig:case_study}, middle, left). This nominal trajectory is inefficient, staying too close to the table. Based on $\theta^*$, the human could provide the improved orange trajectory (Figure \ref{fig:case_study}, middle, left) that is more efficient and doesn't emphasize closeness to the table as much. This improvement reduces our feasible set from $\mathcal{F}_{\text{proxy}}$ to $\mathcal{F}_{\text{improvement}}$:
$$\mathcal{F}_{\text{improvement}} = \{\theta \ : \theta^T\phi(\traj_{\text{improved}}) \ge \theta^T\phi(\traj_R) \quad \ \theta \in \mathcal{F}_{\text{proxy}} \} \,.$$
Figure \ref{fig:case_study} (middle, middle) shows the effect of applying this constraint, shrinking the orange feasible set. The feasible set has shrunk, but not enough to guarantee optimal behavior in all environments. The improvement establishes that closeness to the table should not come at the cost of efficiency. As a result, it removes the red trajectory in environment 3, which greatly traded off efficiency for proximity to the table (Figure \ref{fig:case_study}, middle, right). To further fine tune, we will ask the human to answer a trajectory comparison.

\textbf{Trajectory Comparison.}
The robot presents the human with two trajectories (Figure \ref{fig:case_study} bottom, left, orange and gray) and asks which incurs less cost. The human answers "orange", the trajectory that prioritizes efficiency over distance to the table. This comparison feedback shrinks our feasible set from $\mathcal{F}_{\text{improvement}}$ to $\mathcal{F}_{\text{comparison}}$:
\begin{align*}\mathcal{F}_{\text{comparison}} = \{\theta \in \mathcal{F}_{\text{improvement}} \ : & \ \theta^T\phi(\traj_{\text{orange}}) \ge \theta^T\phi(\traj_{\text{gray}}) \} \,.
\end{align*}
We finally see a very small orange feasible set (Figure \ref{fig:case_study}, bottom, middle). Appropriately, in all four environments now, every $\theta \in \mathcal{F}_{\text{comparison}}$ produces a trajectory $\traj_{\theta}$ s.t. $\phi(\traj_{\theta}) = \phi(\traj_{\theta^*})$. This is illustrated in Figure \ref{fig:case_study} (bottom, right) as only the optimal green trajectory remains in each environment.

Our case study showcases the usefulness of combining types of feedback. A designer might start with their best guess at a reward function, the robot might misbehave in new environments, the designer or even end-user might observe this and intervene to correct or stop the robot, etc. -- over time, the robot should narrow in on what people actually want it to do.

%% file: active.tex
\section{Actively selecting which type of feedback to use}\label{app:active}

\begin{figure*}
    \centering
    \includegraphics[width=3cm]{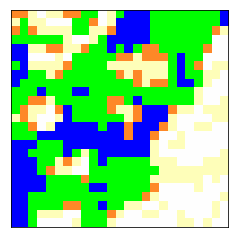}
    \includegraphics[width=3cm]{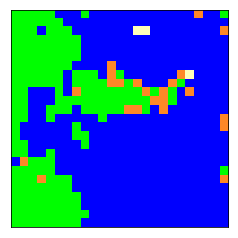}
    \includegraphics[width=3cm]{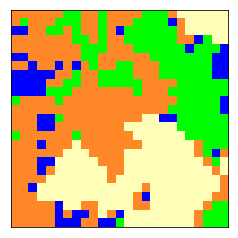}
    \includegraphics[width=3cm]{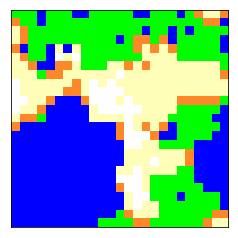}\\
    \includegraphics[width=3cm]{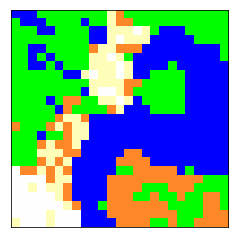}
    \includegraphics[width=3cm]{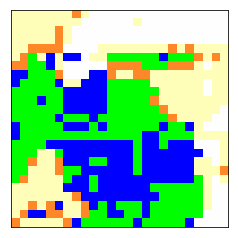}
    \includegraphics[width=3cm]{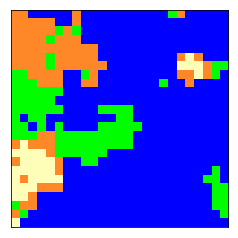}
    \includegraphics[width=3cm]{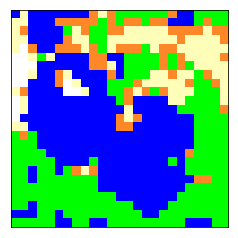}\\
    \caption{Environments used for experiments on active selection of feedback. (Top) These four environments were used during "training". (Bottom) These four environments were held as a test set to measure maximum and average regret.}
    \label{fig:Environments}
\end{figure*}

\begin{figure*}
    \centering
    \includegraphics[width=0.45\textwidth]{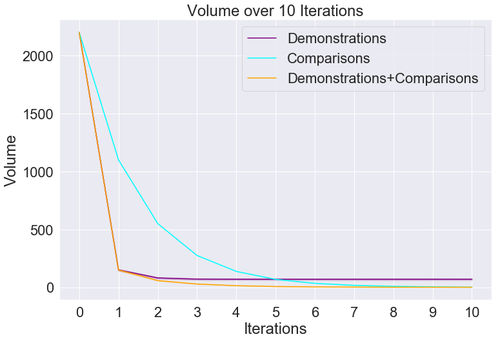}
    \includegraphics[width=0.45\textwidth]{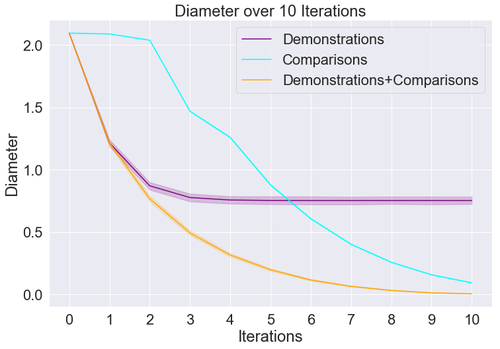} \\
    \includegraphics[width=0.45\textwidth]{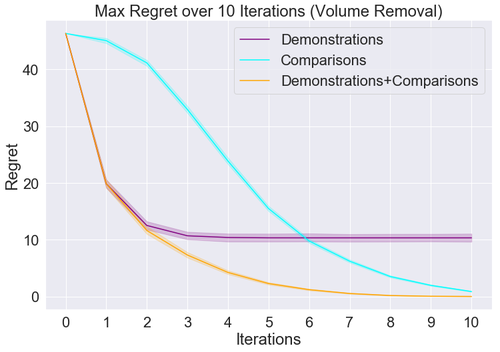}
    \includegraphics[width=0.45\textwidth]{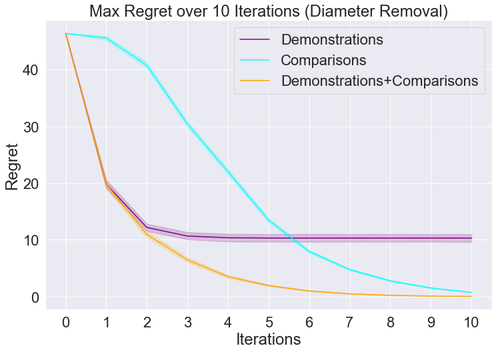} \\
    \includegraphics[width=0.45\textwidth]{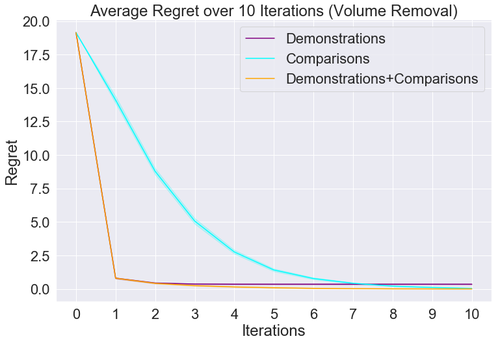}
    \includegraphics[width=0.45\textwidth]{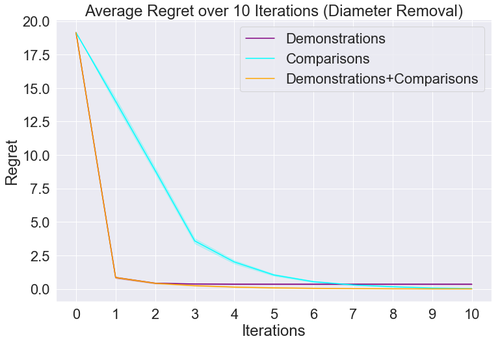}
    \\
    \caption{Statistics computed over 10 iterations of our greedy maximum information gain algorithm. We notice that demonstrations (purple) are initially very information dense but quickly flatten out, whereas comparisons (cyan) obtain more information but less efficiently. We notice that combining the two methods (orange) inherits the positive aspects of both, the efficiency of demonstrations with the precision of comparisons.}
    \label{fig:active-exps}
\end{figure*}

Given we can mix and match types of feedback, we may also wonder what is the best \emph{type} to ask for at each point in time. The probabilistic model defined by reward-rational choice hints at how to select the feedback type -- pick the one that maximizes expected information gain. We point this out, not because using information gain as an active learning metric is a new idea, but because
the ability to use it arises immediately as an application of the formalism. 

Suppose we can select between $n$ types of feedback with choice sets $\choices_1, \dots, \choices_n$ to ask the user for. Let $b_t$ be the robot's belief distribution over rewards at time $t$. The type of feedback $i^*$ that (greedily) maximizes information gain for the next time step is
\begin{align} 
   \nonumber i^* & = \argmax_{i \in [n]} I(\reward_t; \fdbkch_i)   = \ex_{\reward_t, \fdbkch_i} \left [\log \left (\frac{p(\fdbkch_i \mid \reward_t)}{\int_{\reward_t \in \rewards}p(\fdbkch_i \mid \reward_t)b_t(\reward_t)} \right) \right]\,, \label{eq:info-gain}
\end{align}
where $\reward_t \sim B_t$ is distributed according to the robot's current belief, $\fdbkch_i \in \choices_i$ is the random variable corresponding to the user's choice within feedback type $i$, and $p(\fdbkch_i \mid \reward_t)$ is defined according to the human model in Equation \ref{eq:rat-choice2}.

To showcase the benefit of actively selecting feedback types, we run an experiment with demonstrations and comparisons. We measure regret (maximum and expected difference, on holdout environments, in ground truth reward between  1) optimizing with ground truth vs. 2) optimizing with the learned reward). We manipulate whether we have access to demonstrations only, comparisons only, or both, as well as the number of feedback instances queried.

One may initially wonder whether comparisons are necessary, given that demonstrations seem to provide so much information early on. Overall, we observe that demonstrations are optimal early on, when little is known about the reward, while comparisons become optimal later, as a way to fine-tune the reward (\figref{fig:active-exps} shows our results). The observation also serves to validate the approach contributed by \cite{palan2019learning,ibarz2018reward} in the applications of motion planning and Atari game-playing, respectively. Both papers manually define the mixing procedure we found to be optimal: initially train the reward model using human demonstrations, and then fine-tune with comparisons.

\textbf{Experiment Details}
We tested 3 different active learning methods: active querying of demonstrations, active querying of comparisons, and active querying of demonstrations and comparisons, across 8 different gridworld environments depicted in Figure \ref{fig:Environments}. The top 4 environments were used in training while the bottom 4 were held for testing. Each environment $e$ is a 25x25 gridworld MDP with a linear reward function in 3 features: RGB color values of each pixel. We assign each $e$ with 10 different start goal pairs $(s,g)$ from which the algorithms can ask queries. The goal of each algorithm is to efficiently recover a ground truth reward $r^*$ through querying.

Since our rewards are linear in RGB, the feasible reward set $\mathcal{R}$ consists of 3D parameters that weight the value of each feature in the reward function. $\mathcal R$ can be constrained to the surface of the 3D unit sphere since reward functions in MDPs are scale invariant. We uniformly discretize points at the surface of the 3D sphere to approximate $\mathcal{R}$ via $\hat{\mathcal{R}}$. To approximate $\trajs$, we first compute the optimal trajectory under each $r\in \hat{\mathcal{R}}$ to make $\{ \argmax_{\traj}\ r(\traj);\ r\in\hat{\mathcal{R}}\}$. We include trajectories that are not the result of optimizing reward functions by inserting noise into the value function when computing optimal trajectories as above.

\textbf{Demonstrations and Comparisons as Hard Constraints} 
The algorithms recover $r^*$ by narrowing a set of feasible rewards with active queries. We use $\rewards_{i}$ to denote the set of feasible rewards at iteration $i$ of querying. Demonstrations and comparisons shrink the feasible set in the following way:
$$\rewards_{i+1}^{\text{demo}}(\traj_d) = \{r : r(\traj_d) \ge r(\traj) \quad r \in \rewards_i; \quad \forall \traj \in \trajs \}$$
\[ \rewards_{i+1}^{\text{comp}}(\traj_1, \traj_2) =
    \begin{cases} 
      \rewards_{i+1}^{\text{comp}}(\underline{\traj_1}, \traj_2) = \{r: r(\traj_1) \ge r(\traj_2)\quad r\in\rewards_i \} & \traj_1 > \traj_2 \\
      \rewards_{i+1}^{\text{comp}}(\traj_1, \underline{\traj_2}) = \{r: r(\traj_2) \ge r(\traj_1)\quad r\in\rewards_i \} & \traj_2 > \traj_1 \\
   \end{cases}
\]
For our experiments, we performed the following greedy volume removal over possible $(s,g)$ pairs that we specified in each environment.
\[  \rewards_{i+1} =
    \begin{cases}
      \rewards_{i+1}^{\text{demo}}(\traj_d^*)\ & \ V_{\text{demo}} < V_{\text{comp}}\\
  \rewards_{i+1}^{\text{comp}}(\traj_1, \traj_2)\ &\  V_{\text{comp}} < V_{\text{demo}}\\
   \end{cases}
\]
$$V_{\text{comp}} = \min_{(s,g)}\max \ \big\{\rewards_{i+1}^{\text{comp}}(\underline{\traj_1}, \traj_2), \ \rewards_{i+1}^{\text{comp}}(\traj_1, \underline{\traj_2}) \big\}$$
$$V_{\text{demo}} = \min_{(s,g)}\ \mathbb{E}_{r^*\in\rewards_i}\big[|\rewards_{i+1}(\traj_{r^*}^{\text{demo}})|\big]$$

For demonstrations, we look for the $(s,g)$ pair that in expectation produces a demonstrations that leave the smallest feasible set (size of feasible set is volume or diameter described below). For comparisons, we look for the pair of trajectories $(\traj_1, \traj_2)$ that produce the minimum worst-case feasible region remaining. For the method with demonstrations and comparisons, we computed the above 2 metrics and select the feedback type with the smaller feasible region. We run this algorithm for 10 iterations and average our results across 50 different ground truth $r^*$. We plot several statistics for each iteration in Figure \ref{fig:active-exps} including
\[
\begin{cases}
|\rewards_i| & \text{Volume at iteration i}\\
\sup_{r_1, r_2\in\rewards_i} \|r_1-r_2\|_2 & \text{Diameter at iteration i}\\
\max_{e;(s,g); \ r\in\rewards_i}\ r^*(\traj_{r^*}^{(e,s,g)})-r^*(\traj_{r}^{(e,s,g)}) & \text{Max regret at iteration i} \\
\mathbb{E}_{e;(s,g);\ r\in\rewards_i}\ \big[ r^*(\traj_{r^*}^{(e,s,g)})-r^*(\traj_{r}^{(e,s,g)})\big] & \text{Avg regret at iteration i} \\
\end{cases}
\]
where $e$ is a holdout environment and $(s,g)$ is a start-goal pair in the MDP. Each metric is a proxy for how accurate our estimate of $r^*$ is.
We notice that the combination of demonstrations and comparisons achieves lower volume, diameter, max regret, and average regret than demonstrations alone and that it achieves this in fewer iterations than comparisons alone.

%% file: meta.tex
\section{Meta-choice: a new source of information} \label{app:meta}

In Section \ref{sec:disc}, we described a straight-forward way of combining feedback types: treat each individual feedback received as an independent reward rational choice, and update the robot's belief (Equation \ref{eq:prob-combine}). However, the moment we open it up to multiple types of feedback, the person is not stuck with a single type and is actually choosing which type to use. \emph{We propose that this itself is a reward-rational implicit choice, and therefore leaks information about the reward.} We call the choice of feedback ``meta-choice'', and in this section, we formalize it and empirically showcase its potential importance.

\subsection{Formalizing meta-choice}
The assumption of conditional independence that the formulation in (\ref{eq:prob-combine}) uses is natural and makes sense in many settings. For example, during training time, we might control what feedback type we ask the human for. We might start by asking the human for demonstrations, but then move on to other types of feedback, like corrections or comparisons, to get more fine-grained information about the reward function. Since the human is only ever considering one type of feedback at a time, the conditional independence assumption makes sense.

\begin{figure*}[t]
    \centering
    \includegraphics[width=.97\textwidth]{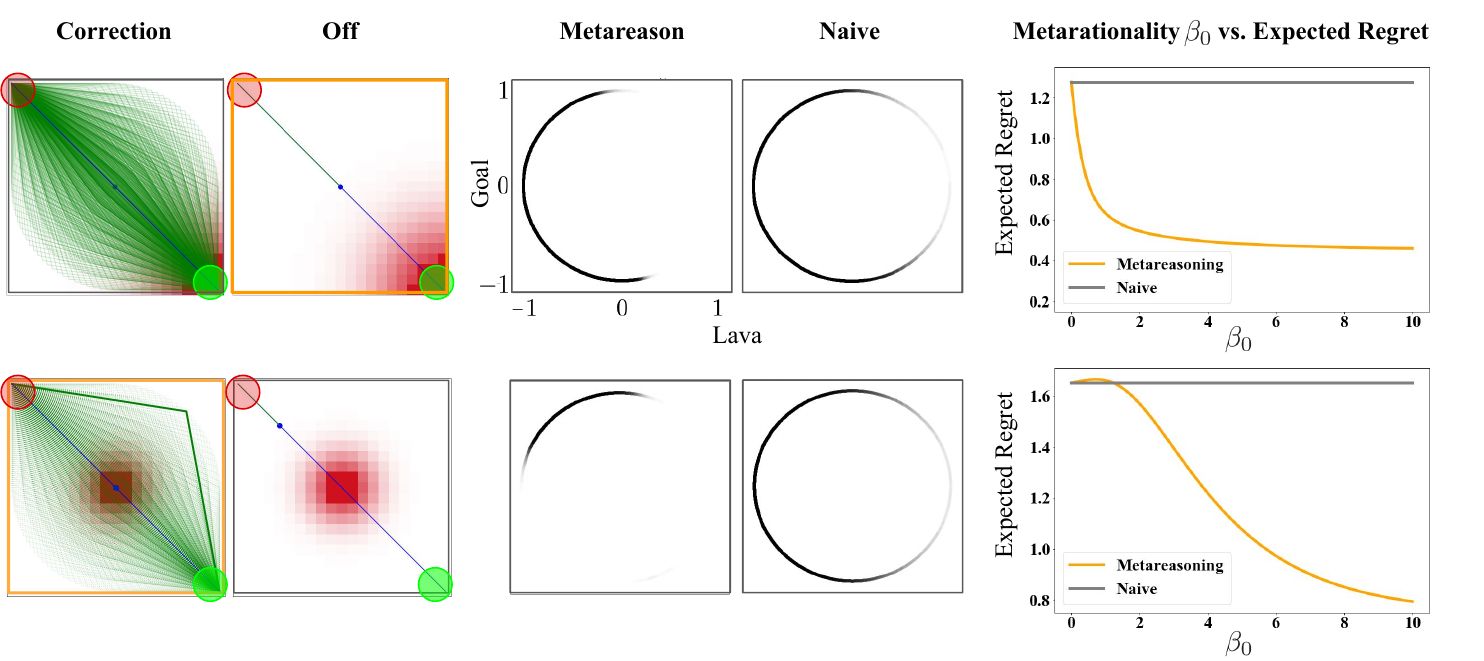}
    \caption{\small{(Left) Environment with designated start (red circle), goal (green circle) and lava area (red tiles). The human can provide a correction (one of the green trajectories) or turn off the robot, forcing the robot to stop at the marked dot. (Middle) Belief distribution over rewards after the human provides  feedback ($\rat_0 = 10.0$). Darker indicates higher probability. The metareasoning model is able to rule out more reward functions than the naive model. (Right) When the human's metareasoning has no signal ($\rat_0 = 0$), then the metareasoning (orange) and naive model (gray) perform equally well. As $\rat_0$ increases, the advantage of the metareasoning model also increases.}}
    \label{fig:metareason_exp}
\end{figure*}

But the assumption breaks when the human has access to multiple types of feedback at once because \emph{the types of feedback the robot can interpret influence what the human does in the first place}.\footnote{We note that this adaptation by the human only applies to types of behavior that the human uses to purposefully communicate with the robot, as opposed to sources of information like initial state.} If the human intervenes and turns the robot off, that means one thing if this were the only feedback type available, and a whole different thing if, say, corrections were available too. In the latter case, we have more information - we know that the user chose to turn the robot off rather than provide a correction. %

Thus, our key insight is that the type of feedback itself leaks information about the reward, and the RRC framework gives us a recipe for formalizing this new source: %
we need to uncover the set of options the human is choosing from. %
The human has two stages of choice: the first is the choice between feedback types, i.e corrections, language, turn-off, etc. and the second is the choice within the chosen feedback type, i.e the specific correction that the human gave. Our formalism can leverage both sources of information by defining a \emph{hierarchy} of reward-rational choice.

Suppose the user has access to $n$ types of feedback with associated choice sets $\choices_1, \dots, \choices_n$, groundings $\map_1, \dots \map_n$, and Boltzmann rationalities $\beta_1, \dots, \beta_n$. For simplicity, we assume deterministic groundings. The set of choice sets $\mathscr{C}_0$ for the first-stage choice is $\{\choices_1, \dots, \choices_n\}$ The grounding $\map_0: \choices \rightarrow f_{\trajs}$ for the first stage choice maps a feedback type $\choices_i$ to the distribution of trajectories defined by the human's behavior and grounding in the second stage:
\begin{align} \label{eq:first-stage-grounding}
    \map_0(C_i)\ & =\ \pr(\traj \mid r,\, \choices_i)  = \sum_{\fdbk_i \in \choices_i : \map_i(\fdbk_i) = \traj} \pr(\fdbk_i \mid r,\, \choices_i) \,,
\end{align}
where, as usual, $\pr(\fdbk_i \mid r,\, \choices_i)$ is given by Equation \ref{eq:rat-choice2}. Instantiating Equation \ref{eq:rat-choice2} to model the first-stage decision as well, results in the following model for the human picking feedback type $\choices_i$:
\begin{align}
\pr(\choices_i \mid r) = \frac{\exp\big( \beta_0\cdot\mathbb{E}_{\traj \sim \map_0(\choices_i)}[r(\traj)] \big)}{\sum_{j \in [n]}\ \exp\big( \beta_0\cdot\mathbb{E}_{\traj \sim \map_0(\choices_j)}[r(\traj)] \big)} \,,
\end{align}
Finally, the probability that the human gives feedback $\fdbkch$ is
\begin{align}
  \pr(\fdbkch \mid\reward) & = \sum_{i} \pr(\fdbkch \mid \reward,\, \choices_i)\cdot \pr (\choices_i \mid \reward)  \\
  & = \sum_{i} \Bigg ( \frac{\exp\big(\beta_i\cdot \reward(\map_i(c^*))\big)}{\sum_{\fdbk \in \choices_i}\ \exp\big(\beta_i\cdot r(\map_i(c))\big)} \cdot \frac{\exp\big( \beta_0\cdot\mathbb{E}_{\traj \sim \map_0(\choices_i)}[r(\traj)] \big)}{\sum_{j \in [n]}\ \exp\big( \beta_0\cdot\mathbb{E}_{\traj \sim \map_0(\choices_j)}[r(\traj)] \big)} \Bigg ) \,. \label{eq:meta}
\end{align}

The first-stage decision can be interpreted as the human \emph{metareasoning} over the best type of feedback. The benefit of modeling the hierarchy is that we can cleanly separate and consider noise at both the level of metareasoning ($\beta_0$) and the level of execution of feedback ($\beta_1, \dots, \beta_n$). Noise at the metareasoning level models the human's imperfection in picking the optimal type of feedback. Noise at the execution level might model the fact that the human has difficulty in physically correcting a heavy and unintuitive robot.\footnote{Although we modeled rationality with respect to the reward $r$ that the robot should optimize, we can easily extend our formalism to capture that the person might trade-off between that and their own effort -- this is especially interesting at this meta-choice level, where one type of feedback might be much more difficult and thus people might want to avoid it unless it is particularly informative.} 

\subsection{Comparing the literal interpretation to meta-choice}
We showcase the potential importance of accounting for the meta-choice in an experiment  in a gridworld setting, in which an agent navigates to a goal state while avoiding lava (Figure \ref{fig:metareason_exp}, left). The reward function is a linear combination of 2 features that encode the goal and lava. The human has access to two channels of feedback: ``off'' and corrections. We simulate the human feedback as choosing between \emph{feedback types} according to Equation \ref{eq:meta}. We manipulate three factors: 1) whether the robot is \emph{naive}, i.e. only accounts for the information \emph{within} the feedback type, or \emph{metareasons}, i.e. accounts for the other feedback types that were available but not chosen; 2) the meta-rationality parameter $\beta_0$ modeling human imperfection in selecting the optimal type of feedback; and 3) the location of the lava, so that the rational meta-choice changes from off to corrections. We measure regret over holdout environments.

Figure \ref{fig:metareason_exp} (left) depicts the possible grounded trajectories for corrections and for off. For the top, off is optimal because all corrections go through lava. For the bottom, the rational meta-choice is to correct. In both cases, we find that meta-reasoning gains the learner more information, as seen in the belief (center). For the top, where the person turns it off, the robot can be more confident that lava is bad. For the bottom, the fact that the person had the off option and did not use it informs the robot about the importance of reaching the goal. This translates into lower regret (right), especially as $\beta_0$ increases and there is more signal in the feedback type choice.

\subsection{What happens when metarationality is misspecified?}
\input{meta-exps}

%% file: meta-exps.tex
In our main metareasoning experiments, we assumed that the simulated human metareasoned with $\rat_0$ and that our algorithm somehow knew this quantity. However, in practice, we will not have access to $\rat_0$. This brings about an interesting question: What are the effects of inference under a misspecified $\rat_0$. What are the effects of overestimating or underestimating the human's rationality?

\begin{figure*}
    \centering
    \includegraphics[width=\textwidth]{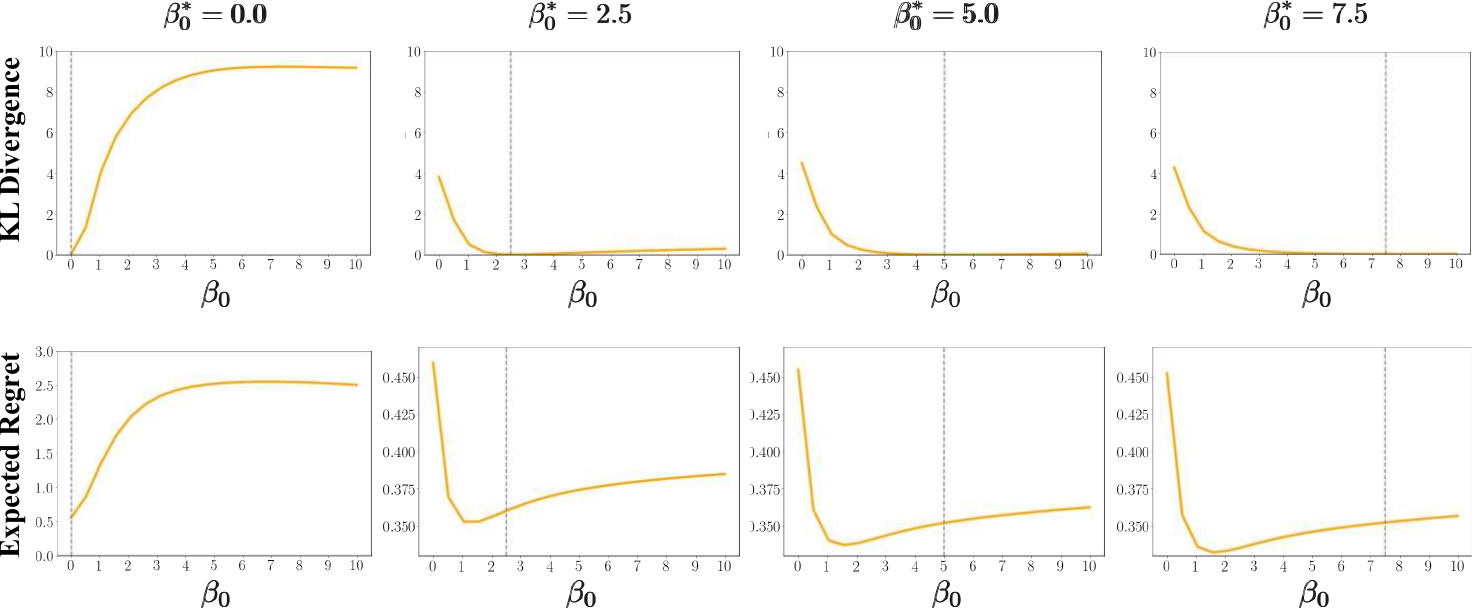}
    \caption{In each plot, the human operates under a true metarationality denoted by $\beta_0^*$. We measure performance drop from misspecification by computing the KL divergence and expected regret of the belief distribution over rewards for robots with misspecified metarationalities $\beta_0 \ \in [0.0, 10.0]$. (Top) The plots display the KL divergence between the true belief with $\beta_0^*$ and various misspecified beliefs. We notice that assuming metareasoning when the human does not metareason (left $\beta_0^* = 0$) results in significant divergence in the belief distribution. (Bottom) The plots show the expected regret for robots that learn, with misspecified $\beta_0$'s, from a human who gives feedback with $\beta_0^*$. As with \emph{KL divergence}, the \emph{expected regret} incurred by assuming metareasoning when the human does not metareason is high. Additionally, we note that a robot learning with $\beta_0=\beta_0^*$ does not necessarily incur minimum expected regret (as mentioned below in the text).}
    \label{fig:misspecification_exp}
\end{figure*}

To test this, we designed an experiment in which our simulated human provided supervision with a fixed ground truth $r^*$ and $\beta^*_0$ while our algorithm performs belief updates with various $\rat_0$ above and below $\beta^*_0$. The first way to measure the extent of misspecification is to measure the KL divergence between the belief induced by $\beta^*_0$ and that induced by $\rat_0$.
$$D_{KL}(P(r\mid c^*)\|Q(r\mid c^*))$$
$$P(r\mid c^*) \propto P(c^*\mid r,\beta^*_0)\cdot P(r)$$
$$Q(r\mid c^*) \propto P(c^*\mid r, \rat_0)\cdot Q(r)$$
Additionally, we wanted to measure the expected regret given a human that provides supervision with rationality $\beta^*_0$ and the algorithm that performs belief updates with rationality $\rat_0$. 
\begin{align*}
\mathbb{E}[\text{Regret} \mid c^*, C_i, r^*, \rat_0] = \sum_{r\in \mathcal{R}}\ (r^*(\phi(r^*))-r^*(\phi(r)))\\
\cdot \pr(r \mid c^*, C_i, \rat_0)
\end{align*}
\begin{align*}
    \mathbb{E}[\text{Regret} \mid \rat_0] = \sum_{r^*\in \mathcal{R}}\sum_{i\in \mathcal{C}_0}\sum_{c^*\in C_i} \ & \mathbb{E}[\text{Regret} \mid c^*, C_i, r^*, \rat_0] \\ & \cdot \pr(C_i \mid r^*, \rat_0) \cdot \pr(c^*\mid C_i,r^*,\rat_0)
\end{align*}
We plot the results in Figure \ref{fig:misspecification_exp} averaged over 50 randomly sampled reward functions and $beta_0\in [0.0, 10.0]$. We notice that when the human does not metareason ($\rat_0^*=0.0$, the KL divergence in the belief distribution update is large. In comparison, with any moderate level of metareasoning $\rat_0^*=2.5, 5.0, 7.5$, the KL divergence is very low. We notice this too in the expected regret. Note that the minimum expected regret is not achieved by $\rat_0 = \rat_0^*$. This is because $\rat_0^*$ is used to compute the frequency at which the human provides each type of feedback as an answer. Simply matching $\rat_0$ with $\rat_0^*$ doesn't guarantee minimum expected regret (the optimal $\rat_0$ for minimizing expected regret is a function of $\rat_0^*$). These experiments suggest that if we detect that the human is poor at metareasoning (low $\rat_0^*$), it is safer to drop the metareasoning assumption. However, if the human is displaying metareasoning, we can leverage this to improve learning.

%% file: main.bbl
\begin{thebibliography}{53}
\providecommand{\natexlab}[1]{#1}
\providecommand{\url}[1]{\texttt{#1}}
\expandafter\ifx\csname urlstyle\endcsname\relax
  \providecommand{\doi}[1]{doi: #1}\else
  \providecommand{\doi}{doi: \begingroup \urlstyle{rm}\Url}\fi

\bibitem[Hadfield-Menell et~al.(2017{\natexlab{a}})Hadfield-Menell, Milli,
  Abbeel, Russell, and Dragan]{hadfield2017inverse}
Dylan Hadfield-Menell, Smitha Milli, Pieter Abbeel, Stuart~J Russell, and Anca
  Dragan.
\newblock Inverse reward design.
\newblock In \emph{Advances in neural information processing systems}, pages
  6765--6774, 2017{\natexlab{a}}.

\bibitem[Ratner et~al.(2018)Ratner, Hadfield-Menell, and
  Dragan]{ratner2018simplifying}
Ellis Ratner, Dylan Hadfield-Menell, and Anca~D Dragan.
\newblock Simplifying reward design through divide-and-conquer.
\newblock \emph{arXiv preprint arXiv:1806.02501}, 2018.

\bibitem[Ng and Russell(2000)]{ng2000algorithms}
Andrew~Y Ng and Stuart~J Russell.
\newblock Algorithms for inverse reinforcement learning.
\newblock In \emph{Icml}, volume~1, page~2, 2000.

\bibitem[Abbeel and Ng(2004)]{abbeel2004apprenticeship}
Pieter Abbeel and Andrew~Y Ng.
\newblock Apprenticeship learning via inverse reinforcement learning.
\newblock In \emph{Proceedings of the twenty-first international conference on
  Machine learning}, page~1. ACM, 2004.

\bibitem[Wirth et~al.(2017)Wirth, Akrour, Neumann, and
  F{\"u}rnkranz]{wirth2017survey}
Christian Wirth, Riad Akrour, Gerhard Neumann, and Johannes F{\"u}rnkranz.
\newblock A survey of preference-based reinforcement learning methods.
\newblock \emph{The Journal of Machine Learning Research}, 18\penalty0
  (1):\penalty0 4945--4990, 2017.

\bibitem[Sadigh et~al.(2017)Sadigh, Dragan, Sastry, and
  Seshia]{dorsa2017active}
Dorsa Sadigh, Anca~D. Dragan, Shankar Sastry, and Sanjit~A Seshia.
\newblock Active preference-based learning of reward functions.
\newblock In \emph{Robotics: Science and Systems (RSS)}, 2017.

\bibitem[Christiano et~al.(2017)Christiano, Leike, Brown, Martic, Legg, and
  Amodei]{christiano2017deep}
Paul~F Christiano, Jan Leike, Tom Brown, Miljan Martic, Shane Legg, and Dario
  Amodei.
\newblock Deep reinforcement learning from human preferences.
\newblock In \emph{Advances in Neural Information Processing Systems}, pages
  4299--4307, 2017.

\bibitem[MacGlashan et~al.(2015)MacGlashan, Babes-Vroman, desJardins, Littman,
  Muresan, Squire, Tellex, Arumugam, and Yang]{macglashan2015grounding}
James MacGlashan, Monica Babes-Vroman, Marie desJardins, Michael~L Littman,
  Smaranda Muresan, Shawn Squire, Stefanie Tellex, Dilip Arumugam, and Lei
  Yang.
\newblock Grounding english commands to reward functions.
\newblock In \emph{Robotics: Science and Systems}, 2015.

\bibitem[Fu et~al.(2019)Fu, Korattikara, Levine, and
  Guadarrama]{fu2019language}
Justin Fu, Anoop Korattikara, Sergey Levine, and Sergio Guadarrama.
\newblock From language to goals: Inverse reinforcement learning for
  vision-based instruction following.
\newblock \emph{arXiv preprint arXiv:1902.07742}, 2019.

\bibitem[Jain et~al.(2015)Jain, Sharma, Joachims, and Saxena]{jain2015learning}
Ashesh Jain, Shikhar Sharma, Thorsten Joachims, and Ashutosh Saxena.
\newblock Learning preferences for manipulation tasks from online coactive
  feedback.
\newblock \emph{The International Journal of Robotics Research}, 34\penalty0
  (10):\penalty0 1296--1313, 2015.

\bibitem[Bajcsy et~al.(2017)Bajcsy, Losey, O’Malley, and
  Dragan]{bajcsy2017learning}
Andrea Bajcsy, Dylan~P Losey, Marcia~K O’Malley, and Anca~D Dragan.
\newblock Learning robot objectives from physical human interaction.
\newblock \emph{Conference on Robot Learning (CoRL)}, 2017.

\bibitem[Shah et~al.(2019)Shah, Krasheninnikov, Alexander, Abbeel, and
  Dragan]{shah2019preferences}
Rohin Shah, Dmitrii Krasheninnikov, Jordan Alexander, Pieter Abbeel, and Anca
  Dragan.
\newblock Preferences implicit in the state of the world.
\newblock In \emph{ICLR}, 2019.

\bibitem[Baker et~al.(2009)Baker, Saxe, and Tenenbaum]{baker2009action}
Chris~L Baker, Rebecca Saxe, and Joshua~B Tenenbaum.
\newblock Action understanding as inverse planning.
\newblock \emph{Cognition}, 113\penalty0 (3):\penalty0 329--349, 2009.

\bibitem[Goodman et~al.(2009)Goodman, Baker, and Tenenbaum]{goodman2009cause}
Noah~D Goodman, Chris~L Baker, and Joshua~B Tenenbaum.
\newblock Cause and intent: Social reasoning in causal learning.
\newblock In \emph{Proceedings of the 31st annual conference of the cognitive
  science society}, pages 2759--2764. Citeseer, 2009.

\bibitem[Goodman and Stuhlm{\"u}ller(2013)]{goodman2013knowledge}
Noah~D Goodman and Andreas Stuhlm{\"u}ller.
\newblock Knowledge and implicature: Modeling language understanding as social
  cognition.
\newblock \emph{Topics in cognitive science}, 5\penalty0 (1):\penalty0
  173--184, 2013.

\bibitem[Bradley and Terry(1952)]{bradley1952rank}
Ralph~Allan Bradley and Milton~E Terry.
\newblock Rank analysis of incomplete block designs: I. the method of paired
  comparisons.
\newblock \emph{Biometrika}, 39\penalty0 (3/4):\penalty0 324--345, 1952.

\bibitem[Luce(1959)]{luce1959individual}
R~Duncan Luce.
\newblock \emph{Individual choice behavior: A theoretical analysis}.
\newblock Wiley, 1959.

\bibitem[Plackett(1975)]{plackett1975analysis}
Robin~L Plackett.
\newblock The analysis of permutations.
\newblock \emph{Journal of the Royal Statistical Society: Series C (Applied
  Statistics)}, 24\penalty0 (2):\penalty0 193--202, 1975.

\bibitem[Ziebart et~al.(2008)Ziebart, Maas, Bagnell, and
  Dey]{ziebart2008maximum}
Brian~D Ziebart, Andrew~L Maas, J~Andrew Bagnell, and Anind~K Dey.
\newblock Maximum entropy inverse reinforcement learning.
\newblock In \emph{Aaai}, volume~8, pages 1433--1438. Chicago, IL, USA, 2008.

\bibitem[Ramachandran and Amir(2007)]{ramachandran2007bayesian}
Deepak Ramachandran and Eyal Amir.
\newblock Bayesian inverse reinforcement learning.
\newblock In \emph{IJCAI}, volume~7, pages 2586--2591, 2007.

\bibitem[Finn et~al.(2016)Finn, Levine, and Abbeel]{finn2016guided}
Chelsea Finn, Sergey Levine, and Pieter Abbeel.
\newblock Guided cost learning: Deep inverse optimal control via policy
  optimization.
\newblock In \emph{International Conference on Machine Learning}, pages 49--58,
  2016.

\bibitem[Bloem and Bambos(2014)]{bloem2014infinite}
Michael Bloem and Nicholas Bambos.
\newblock Infinite time horizon maximum causal entropy inverse reinforcement
  learning.
\newblock In \emph{53rd IEEE Conference on Decision and Control}, pages
  4911--4916. IEEE, 2014.

\bibitem[Dragan et~al.(2013)Dragan, Lee, and Srinivasa]{dragan2013legibility}
Anca~D Dragan, Kenton~CT Lee, and Siddhartha~S Srinivasa.
\newblock Legibility and predictability of robot motion.
\newblock In \emph{Proceedings of the 8th ACM/IEEE international conference on
  Human-robot interaction}, pages 301--308. IEEE Press, 2013.

\bibitem[Blei et~al.(2017)Blei, Kucukelbir, and McAuliffe]{blei2017variational}
David~M Blei, Alp Kucukelbir, and Jon~D McAuliffe.
\newblock Variational inference: A review for statisticians.
\newblock \emph{Journal of the American Statistical Association}, 112\penalty0
  (518):\penalty0 859--877, 2017.

\bibitem[Ibarz et~al.(2018)Ibarz, Leike, Pohlen, Irving, Legg, and
  Amodei]{ibarz2018reward}
Borja Ibarz, Jan Leike, Tobias Pohlen, Geoffrey Irving, Shane Legg, and Dario
  Amodei.
\newblock Reward learning from human preferences and demonstrations in {A}tari.
\newblock In \emph{Advances in Neural Information Processing Systems
  (NeurIPS)}, 2018.

\bibitem[Ratliff et~al.(2006)Ratliff, Bagnell, and
  Zinkevich]{ratliff2006maximum}
Nathan~D Ratliff, J~Andrew Bagnell, and Martin~A Zinkevich.
\newblock Maximum margin planning.
\newblock In \emph{Proceedings of the 23rd international conference on Machine
  learning}, pages 729--736. ACM, 2006.

\bibitem[Hadfield-Menell et~al.(2017{\natexlab{b}})Hadfield-Menell, Dragan,
  Abbeel, and Russell]{hadfield2017off}
Dylan Hadfield-Menell, Anca Dragan, Pieter Abbeel, and Stuart Russell.
\newblock The off-switch game.
\newblock In \emph{Workshops at the Thirty-First AAAI Conference on Artificial
  Intelligence}, 2017{\natexlab{b}}.

\bibitem[Matuszek et~al.(2012)Matuszek, FitzGerald, Zettlemoyer, Bo, and
  Fox]{matuszek2012joint}
Cynthia Matuszek, Nicholas FitzGerald, Luke Zettlemoyer, Liefeng Bo, and Dieter
  Fox.
\newblock A joint model of language and perception for grounded attribute
  learning.
\newblock In \emph{International Conference on Machine Learning (ICML)}, 2012.

\bibitem[Griffith et~al.(2013)Griffith, Subramanian, Scholz, Isbell, and
  Thomaz]{griffith2013policy}
Shane Griffith, Kaushik Subramanian, Jonathan Scholz, Charles~L Isbell, and
  Andrea~L Thomaz.
\newblock Policy shaping: Integrating human feedback with reinforcement
  learning.
\newblock In \emph{Advances in neural information processing systems}, pages
  2625--2633, 2013.

\bibitem[Wirth et~al.(2016)Wirth, F{\"u}rnkranz, and Neumann]{wirth2016model}
Christian Wirth, Johannes F{\"u}rnkranz, and Gerhard Neumann.
\newblock Model-free preference-based reinforcement learning.
\newblock In \emph{Thirtieth AAAI Conference on Artificial Intelligence}, 2016.

\bibitem[Wilson et~al.(2012)Wilson, Fern, and Tadepalli]{wilson2012bayesian}
Aaron Wilson, Alan Fern, and Prasad Tadepalli.
\newblock A bayesian approach for policy learning from trajectory preference
  queries.
\newblock In \emph{Advances in neural information processing systems}, pages
  1133--1141, 2012.

\bibitem[Busa-Fekete et~al.(2013)Busa-Fekete, Sz{\"o}r{\'e}nyi, Weng, Cheng,
  and H{\"u}llermeier]{busa2013preference}
R{\'o}bert Busa-Fekete, Bal{\'a}zs Sz{\"o}r{\'e}nyi, Paul Weng, Weiwei Cheng,
  and Eyke H{\"u}llermeier.
\newblock Preference-based evolutionary direct policy search.
\newblock In \emph{ICRA Workshop on Autonomous Learning}, 2013.

\bibitem[Akrour et~al.(2011)Akrour, Schoenauer, and
  Sebag]{akrour2011preference}
Riad Akrour, Marc Schoenauer, and Michele Sebag.
\newblock Preference-based policy learning.
\newblock In \emph{Joint European Conference on Machine Learning and Knowledge
  Discovery in Databases}, pages 12--27. Springer, 2011.

\bibitem[Wirth and F{\"u}rnkranz(2014)]{wirth2014learning}
Christian Wirth and Johannes F{\"u}rnkranz.
\newblock On learning from game annotations.
\newblock \emph{IEEE Transactions on Computational Intelligence and AI in
  Games}, 7\penalty0 (3):\penalty0 304--316, 2014.

\bibitem[Holladay et~al.(2016)Holladay, Javdani, Dragan, and
  Srinivasa]{holladay2016active}
Rachel Holladay, Shervin Javdani, Anca Dragan, and Siddhartha Srinivasa.
\newblock Active comparison based learning incorporating user uncertainty and
  noise.
\newblock In \emph{RSS Workshop on Model Learning for Human-Robot
  Communication}, 2016.

\bibitem[Ho and Ermon(2016)]{ho2016generative}
Jonathan Ho and Stefano Ermon.
\newblock Generative adversarial imitation learning.
\newblock In \emph{Advances in Neural Information Processing Systems}, pages
  4565--4573, 2016.

\bibitem[Levine and Koltun(2012)]{levine2012continuous}
Sergey Levine and Vladlen Koltun.
\newblock Continuous inverse optimal control with locally optimal examples.
\newblock \emph{arXiv preprint arXiv:1206.4617}, 2012.

\bibitem[Dragan and Srinivasa(2012)]{dragan2012formalizing}
Anca~D Dragan and Siddhartha~S Srinivasa.
\newblock \emph{Formalizing assistive teleoperation}.
\newblock MIT Press, July, 2012.

\bibitem[Haddadin et~al.(2008)Haddadin, Albu-Schaffer, De~Luca, and
  Hirzinger]{haddadin2008collision}
Sami Haddadin, Alin Albu-Schaffer, Alessandro De~Luca, and Gerd Hirzinger.
\newblock Collision detection and reaction: A contribution to safe physical
  human-robot interaction.
\newblock In \emph{2008 IEEE/RSJ International Conference on Intelligent Robots
  and Systems}, pages 3356--3363. IEEE, 2008.

\bibitem[Hogan(1985)]{hogan1985impedance}
Neville Hogan.
\newblock Impedance control: An approach to manipulation: Part
  ii—implementation.
\newblock 1985.

\bibitem[Losey and O’Malley(2017)]{losey2017trajectory}
Dylan~P Losey and Marcia~K O’Malley.
\newblock Trajectory deformations from physical human--robot interaction.
\newblock \emph{IEEE Transactions on Robotics}, 34\penalty0 (1):\penalty0
  126--138, 2017.

\bibitem[Losey and O'Malley(2018)]{losey2018including}
Dylan~P Losey and Marcia~K O'Malley.
\newblock Including uncertainty when learning from human corrections.
\newblock \emph{arXiv preprint arXiv:1806.02454}, 2018.

\bibitem[Fried et~al.(2018{\natexlab{a}})Fried, Andreas, and
  Klein]{fried2018unified}
Daniel Fried, Jacob Andreas, and Dan Klein.
\newblock Unified pragmatic models for generating and following instructions.
\newblock \emph{NAACL}, 2018{\natexlab{a}}.

\bibitem[Fried et~al.(2018{\natexlab{b}})Fried, Hu, Cirik, Rohrbach, Andreas,
  Morency, Berg-Kirkpatrick, Saenko, Klein, and Darrell]{fried2018speaker}
Daniel Fried, Ronghang Hu, Volkan Cirik, Anna Rohrbach, Jacob Andreas,
  Louis-Philippe Morency, Taylor Berg-Kirkpatrick, Kate Saenko, Dan Klein, and
  Trevor Darrell.
\newblock Speaker-follower models for vision-and-language navigation.
\newblock In \emph{Advances in Neural Information Processing Systems}, pages
  3314--3325, 2018{\natexlab{b}}.

\bibitem[Tellex et~al.(2011)Tellex, Kollar, Dickerson, Walter, Banerjee,
  Teller, and Roy]{tellex2011understanding}
Stefanie Tellex, Thomas Kollar, Steven Dickerson, Matthew~R Walter, Ashis~Gopal
  Banerjee, Seth Teller, and Nicholas Roy.
\newblock Understanding natural language commands for robotic navigation and
  mobile manipulation.
\newblock In \emph{Twenty-Fifth AAAI Conference on Artificial Intelligence},
  2011.

\bibitem[Mindermann et~al.(2018)Mindermann, Shah, Gleave, and
  Hadfield-Menell]{activeird:2018}
Sören Mindermann, Rohin Shah, Adam Gleave, and Dylan Hadfield-Menell.
\newblock Active inverse reward design.
\newblock In \emph{Proceedings of the 1st Workshop on Goal Specifications for
  Reinforcement Learning}, 2018.

\bibitem[Loftin et~al.(2014)Loftin, MacGlashan, Peng, Taylor, Littman, Huang,
  and Roberts]{loftin2014strategy}
Robert~Tyler Loftin, James MacGlashan, Bei Peng, Matthew~E Taylor, Michael~L
  Littman, Jeff Huang, and David~L Roberts.
\newblock A strategy-aware technique for learning behaviors from discrete human
  feedback.
\newblock In \emph{AAAI Conference on Artificial Intelligence}, 2014.

\bibitem[MacGlashan et~al.(2017)MacGlashan, Ho, Loftin, Peng, Wang, Roberts,
  Taylor, and Littman]{macglashan2017interactive}
James MacGlashan, Mark~K Ho, Robert Loftin, Bei Peng, Guan Wang, David~L
  Roberts, Matthew~E Taylor, and Michael~L Littman.
\newblock Interactive learning from policy-dependent human feedback.
\newblock In \emph{Proceedings of the 34th International Conference on Machine
  Learning-Volume 70}, pages 2285--2294. JMLR.org, 2017.

\bibitem[Palan et~al.(2019)Palan, Landolfi, Shevchuk, and
  Sadigh]{palan2019learning}
Malayandi Palan, Nicholas~C Landolfi, Gleb Shevchuk, and Dorsa Sadigh.
\newblock Learning reward functions by integrating human demonstrations and
  preferences.
\newblock In \emph{RSS}, 2019.

\bibitem[Ortega and Braun(2013)]{ortega2013thermodynamics}
Pedro~A Ortega and Daniel~A Braun.
\newblock Thermodynamics as a theory of decision-making with
  information-processing costs.
\newblock \emph{Proceedings of the Royal Society A: Mathematical, Physical and
  Engineering Sciences}, 469\penalty0 (2153):\penalty0 20120683, 2013.

\bibitem[Simon(1956)]{simon1956rational}
Herbert~A Simon.
\newblock Rational choice and the structure of the environment.
\newblock \emph{Psychological review}, 63\penalty0 (2):\penalty0 129, 1956.

\bibitem[Jaynes(1957)]{jaynes1957information}
Edwin~T Jaynes.
\newblock Information theory and statistical mechanics.
\newblock \emph{Physical review}, 106\penalty0 (4):\penalty0 620, 1957.

\bibitem[Schulman et~al.(2014)Schulman, Duan, Ho, Lee, Awwal, Bradlow, Pan,
  Patil, Goldberg, and Abbeel]{Schulmanetal_IJRR2014}
John Schulman, Yan Duan, Jonathan Ho, Alex Lee, Ibrahim Awwal, Henry Bradlow,
  Jia Pan, Sachin Patil, Ken Goldberg, and Pieter Abbeel.
\newblock Motion planning with sequential convex optimization and convex
  collision checking.
\newblock \emph{International Journal of Robotics Research (IJRR)}, 2014.

\end{thebibliography}
